\documentclass[runningheads]{llncs}
\usepackage[T1]{fontenc}
\usepackage{graphicx}
\usepackage{amsfonts}
\usepackage{amsmath}
\usepackage{algorithm}
\usepackage{algpseudocode}
\usepackage{subfig}
\usepackage{hyperref}
\usepackage{gensymb}
\usepackage{multirow}
\usepackage{array}
\usepackage{booktabs}
\usepackage{makecell}
\usepackage{bm}
\usepackage{adjustbox}
\usepackage[title]{appendix}
\begin{document}
\sloppy
\title{China Regional 3km Downscaling Based on Residual Corrective Diffusion Model}
\titlerunning{Residual Corrective Diffusion-Based 3km Downscaling}
%
%
\author{Honglu Sun\inst{1,2} \and
Hao Jing\inst{1,2}\thanks{Corresponding author: jingh@cma.gov.cn } \and
Zhixiang Dai\inst{3} \and 
Sa Xiao\inst{1,2} \and 
Wei Xue\inst{4} \and
Jian Sun\inst{1,2} \and
Qifeng Lu\inst{1,2}
}
\authorrunning{Honglu Sun et al.}
%
\institute{State Key Laboratory of Severe Weather Meteorological Science and Technology (LaSW), Beijing, China
\and
CMA Earth System Modeling and Prediction Centre (CEMC), Beijing, China
\and
NVIDIA
\and
Tsinghua University, Beijing, China
}
\maketitle              
\begin{abstract}
A fundamental challenge in numerical weather prediction is to efficiently produce high-resolution forecasts.
A common solution is applying downscaling methods, which include dynamical downscaling and statistical downscaling, to the outputs of global models, such as numerical weather prediction models and data-driven models.
This work focuses on statistical downscaling, which establishes statistical relationships between low-resolution and high-resolution historical data using statistical models.
Deep learning has emerged as a powerful tool for this task, giving rise to various high-performance super-resolution models, which can be directly applied for downscaling, such as diffusion models and Generative Adversarial Networks.
This work relies on a diffusion-based downscaling framework named CorrDiff.
In contrast to the original work of CorrDiff, the region considered in this work is nearly 40 times larger, and we not only consider surface variables as in the original work, but also encounter high-level variables (six pressure levels) as target downscaling variables. In addition, a global residual connection is added to improve accuracy.
In order to generate the 3km forecasts for the China region, we apply our trained models to the 25km global grid forecasts of CMA-GFS, an operational global model of the China Meteorological Administration (CMA), and SFF, a data-driven deep learning-based weather model developed from Spherical Fourier Neural Operators (SFNO). 
CMA-MESO, a high-resolution regional model, is chosen as the baseline model.
The experimental results demonstrate that the forecasts downscaled by our method generally outperform the direct forecasts of CMA-MESO in terms of MAE for the target variables.
Our forecasts of radar composite reflectivity show that CorrDiff, as a generative model, can generate fine-scale details that lead to more realistic predictions compared to the corresponding deterministic regression models.

\end{abstract}
\section{Introduction}
Gridded meteorological forecasts are important in various fields such as transportation, energy sector, agriculture, and scientific research.
In particular, high spatial resolution forecasts are crucial for local studies and risk assessment.
Traditionally, global gridded forecasts are obtained from numerical weather prediction models.
Generating km-resolution forecasts is challenging for numerical weather prediction models due to the limitation of computation time.
Currently, there are also data-driven deep learning models that generate global forecasts \cite{bi2022pangu,lam2023learning,pathak2022fourcastnet,chen2023fuxi}.
However, up to now, there are few global high-resolution gridded reanalysis data that can be used to train such data-driven models.
Instead of getting global high-resolution forecasts, a practical alternative is to get regional high-resolution forecasts from the low-resolution output of a global model by downscaling.

Downscaling methods can be categorized into three types: dynamical downscaling, statistical downscaling, and combined methods.
Dynamical downscaling, similar to numerical weather prediction models, is based on a set of atmospheric dynamical equations. It derives high-resolution forecasts by solving these equations using initial and lateral boundary conditions provided by global models. 
Statistical downscaling establishes statistical relationships between low-resolution variables of global models and high-resolution variables using historical data. These relationships are then applied for future forecasting.
Compared to dynamical downscaling, statistical downscaling offers advantages, including simpler implementation, lower computational cost, and potentially higher accuracy. 

Many machine learning models have been applied for statistical downscaling, such as multiple linear regression \cite{schoof2001downscaling}, support vector machine \cite{chen2010downscaling}, random forest \cite{davy2010statistical}, and artificial neural networks \cite{laddimath2019artificial}.
Among these models, artificial neural networks, which have evolved into deep learning \cite{sun2024deep}, are likely the most promising approach.

The fast development in deep learning over the past decade has led to numerous active research areas, including super-resolution in computer vision.
Various high-performance super-resolution models have been proposed, such as Generative Adversarial Networks (GANs) and diffusion models. 
Super-resolution is similarity to downscaling: the input for both is a low-resolution grid, and the output is a high-resolution grid.
However, there is also a difference between super-resolution and downscaling: the goal of super-resolution is to generate visually realistic images, while meteorological downscaling must ensure accuracy and physical consistency.
Given the similarity between super-resolution and downscaling, many researchers have investigated the application of such super-resolution models to downscaling \cite{watt2024generative,addison2022machine}.
In parallel, there are also works that developed specific neural network structures for downscaling \cite{wu2024gsdnet}.

This work investigates a diffusion-based downscaling model named Corrective Diffusion (CorrDiff) \cite{mardani2025residual}.
CorrDiff is a two-step approach that includes the training of a regression model and the training of a diffusion model to improve the predictions of the regression model.
In \cite{mardani2025residual}, CorrDiff is applied to the Taiwan region, the resolution of the inputs is 25km and the resolution of the outputs is 2km. The size of the 2km high-resolution grid is 448 $\times$ 448. In this study, we apply CorrDiff on the China region, the resolutions of the inputs and the outputs are 25km and 3km respectively. The size of our 3km high-resolution grid is 1600 $\times$ 2400, which is nearly 20 times the size in \cite{mardani2025residual}.
Our models are trained on reanalysis data, including 25km ECMWF Reanalysis v5 (ERA5) \cite{hersbach2020era5} (as low-resolution inputs of the downscaling models) and 3km reanalysis data (as high-resolution labels) that are produced by China Meteorological Administration Regional Reanalysis Atmospheric System (CMA-RRA).
In contrast to \cite{mardani2025residual}, which focuses mainly on surface variables, in this work multiple variables are considered for downscaling, including surface variables and variables at six pressure levels.
The prediction of radar composite reflectivity is also investigated.
Different combinations of input and output variables are examined in order to understand the intervariable dependencies in the downscaling task.
By connecting our downscaling models to global forecasts, 3km regional forecasts for the China region are obtained, which are evaluated through comparison with CMA-MESO.
CMA-MESO is a high-resolution regional numerical weather prediction model of the China Meteorological Administration (CMA) that generates 3km and 1km resolution forecasts of the China region.
Two global forecasts are considered: CMA-GFS, which is an operational global model of CMA, and SFF, which is a deep learning-based weather model.
The experimental results demonstrate that, for the mean absolute error (MAE), our forecasts generally outperform those of CMA-MESO for the target variables.
Our assessment of the uncertainty estimated by CorrDiff shows that, for any downscaled variable, the uncertainty is correlated with the accuracy.
For the prediction of radar reflectivity, our results show that deterministic models usually have a severe over-smoothing problem, while CorrDiff can generate realistic small-scale features that are similar to reanalysis data.

This paper is organized as follows.
Section~\ref{problem} formulates the problem.
Section~\ref{model} introduces the CorrDiff framework.
Section~\ref{setup} presents our baseline model (CMA-MESO), the high-resolution reanalysis data used for training, and our training setup.
Section~\ref{result} presents the experimental results:
it begins with an evaluation on the validation data, covering prediction errors and the properties of the uncertainty estimated by our models; 
it then assesses the 3km forecasts generated by applying our models to the forecasts of global models (CMA-GFS and SFF). 
Section~\ref{conclusion} draws conclusions and discusses future directions.

\section{Problem Formulation}
\label{problem}
$\{x_{i} \in \mathbb{R}^{c_{in} \times p \times q} \mid i \in \{1,  ..., N \} \}$  and $\{ y_{i} \in \mathbb{R}^{c_{out} \times m \times n} \mid i \in \{1,...,N \} \}$ represent 25km and 3km resolution meteorological data over the China region, respectively.
$c_{in}$ ($c_{out}$) is the number of variables of the 25km (3km) data.
$p$ and $q$ ($m$ and $n$) represent the number of grid points in the meridional and zonal directions respectively of 25km (3km) data. $N$ is the number of training samples.
In this work, $p=192$, $q=288$, $m=1600$, and $n=2400$. The 25km data correspond to the region 12.25-60\degree N, 70-141.75\degree E and the 3km data correspond to the region 12.13-60.1\degree N, 70-141.97\degree E. Different combinations of $c_{in}$ and $c_{out}$ are investigated, which will be discussed in Section~\ref{training_setup}.
The objective is to train a deep learning model to predict $y_{i}$ from $x_{i}$.

In the case of using a regression model (in this paper, regression models refer to deterministic deep learning models, such as UNet), first, the input $x_i$ is interpolated onto the 3km resolution grid by bilinear interpolation, which is denoted as $bilinear(x_i)$, then $bilinear(x_i)$ is fed into a deep learning model that can be represented by a function $f: \mathbb{R}^{c_{in} \times p \times q} \mapsto \mathbb{R}^{c_{out} \times m \times n}$.
$f(bilinear(x_i))$ is the prediction of the 3km grid that corresponds to $x_i$.
The training of this deep learning model is to find parameters of $f$ that minimize a loss function quantifying the distance between $f(bilinear(x_{i}))$ and $y_i$ for $i \in \{1,...,N \}$.

In this work, we also use diffusion models. Generally, a diffusion model involves two processes: noising and denoising. Noising is the process that progressively adds noise to a clean image until the image becomes random noise. Denoising is the process that transforms random noise into a structured image. Basically, the training of a diffusion model can be considered as the training of a denoising model.
In the case of using diffusion models for downscaling, the low-resolution data are also fed into the diffusion models. Such models are also known as conditional diffusion models.
The overall inference process of a conditional diffusion model for downscaling can be represented by a function $f:\mathbb{R}^{c_{in} \times p \times q},E \mapsto \mathbb{R}^{c_{out} \times m \times n}$, which is a conditional sampling function of the probability of the 3km resolution data given a 25km resolution input. $E$ represents the set of random noise that follows a certain distribution (which normally follows the normal distribution). 
Note that this function $f$ is highly complex because the denoising process contains multiple steps that progressively decrease the noise level.
For more details on diffusion models, see \cite{ho2020denoising}.

\section{Corrective Diffusion Model}
\label{model}
This section briefly introduces the Corrective Diffusion Model (CorrDiff), for more details see \cite{mardani2025residual}.
The code implementation is based on PhysicsNeMo (https://github.com/NVIDIA/physicsnemo).
The training of CorrDiff has two steps.
In the first step, we train a regression model (UNet), denoted as $f$, to minimize a loss function between its predictions $f(bilinear(x_{i}))$ and the targets $y_i$ for all $i \in \{1,...,N \}$.
In the second step, we train an Elucidated Diffusion Model (EDM) \cite{karras2022elucidating} to correct the predictions made by $f$.
We denote the overall inference process of EDM by a function $g$.
After training, the CorrDiff prediction $\hat{y}_i$ for the target $y_i$ is given by $\hat{y}_i = g(bilinear(x_{i}),f(bilinear(x_{i})),\epsilon)+f(bilinear(x_i))$, where $\epsilon$ is a random noise sampled from a standard normal distribution.
By sampling multiple random noises, a CorrDiff model generates an ensemble of predictions, thereby producing a probabilistic prediction of $y_i$.

Following \cite{mardani2025residual}, we use a specific architecture of UNet \cite{karras2022elucidating} for both the regression (first step) and diffusion (second step) networks.
This architecture has 6 encoder and decoder layers, and it incorporates attention mechanisms and residual connections within its structure.

\begin{figure}[!htbp]
\centering
\includegraphics[width=1\linewidth]{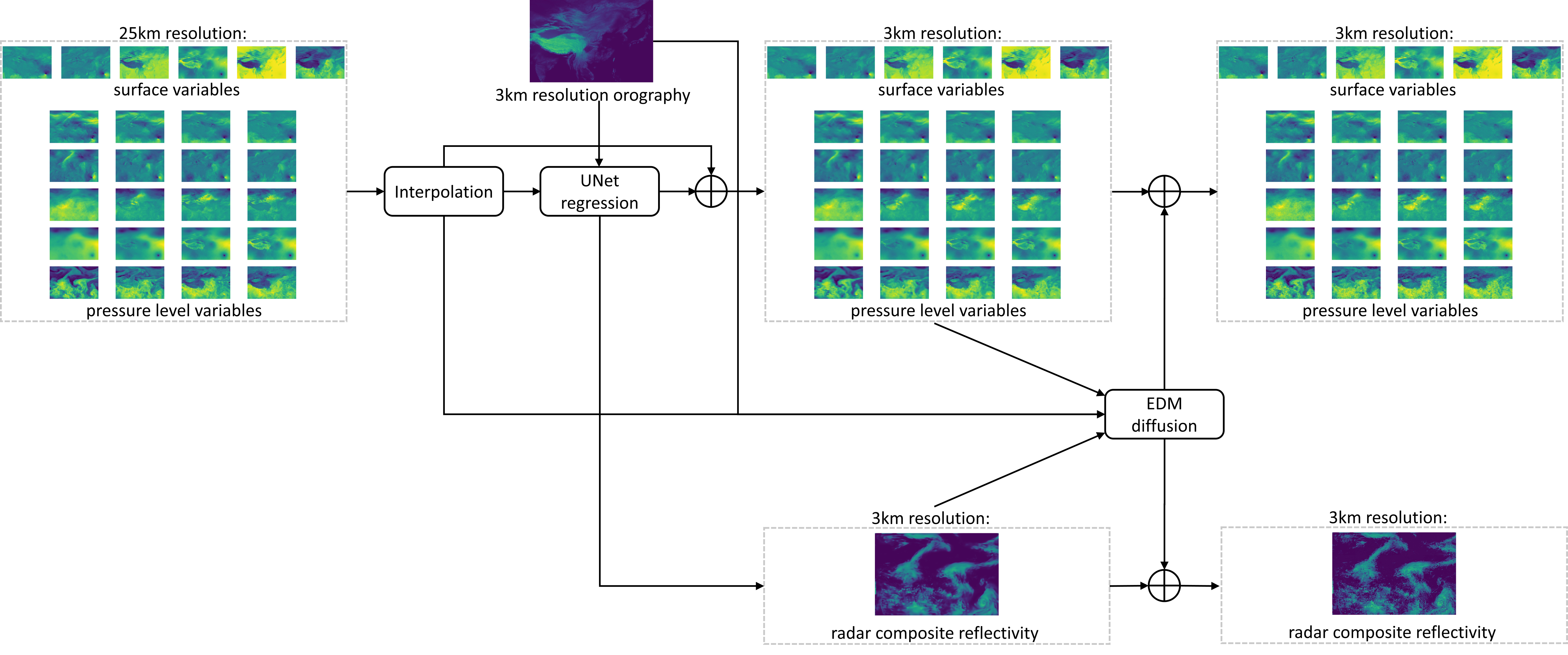}
\caption{Illustration of the overall architecture of one of our trained CorrDiff models.} \label{fig_corrdiff_illus}
\end{figure}

Fig. \ref{fig_corrdiff_illus} shows the overall architecture of one of our trained CorrDiff models.
Our regression networks differ slightly from the models in \cite{mardani2025residual}: for the downscaling of a variable, assuming that $x_i^{u}$ and $y_i^{v}$ correspond to this variable (where $x_i^{u} \in \mathbb{R}^{p \times q} $ is a channel of $x_i$,  and $y_i^{v} \in \mathbb{R}^{m \times n} $ is a channel of $y_i$), instead of directly predicting $y_{i}^{v}$, we predict the residual $y_{i}^{v} - bilinear(x_i^{u})$, because our experimental results show that this residual learning strategy could accelerate convergence speed and improve accuracy for the downscaling task.
In contrast to \cite{mardani2025residual} which only considers the downscaling of surface variables, this work also investigates the downscaling of pressure level variables. An additional input (3km resolution orography) is included in our models as well. 
The implementation of our models is based on the code provided in \cite{mardani2025residual}.

As stated in \cite{mardani2025residual}, the development of CorrDiff was motivated by the limitations of using conditional diffusion models for downscaling.
It is assumed that there is a significant distribution shift between low-resolution and high-resolution data that hinders the learning process.
To avoid this problem, the generation is decomposed into two steps. The first step aims to predict the conditional mean using regression (UNet), and the second step learns a correction using a diffusion model.

\section{Experimental Setup}
\label{setup}
\subsection{Baseline Model: CMA-MESO}
 The China Meteorological Administration Mesoscale (CMA-MESO) model is a high-resolution regional model developed by the China Meteorological Administration Earth System
 Modeling and Prediction Centre (CEMC)\cite{liping2022key}. 
 The model is based on a fully compressible non-hydrostatic dynamical core that uses a semi-implicit semi-Lagrangian discretization scheme. 
 It covers the region of 10–60.1\degree N, 70–145\degree E.
 Currently, CMA-MESO has two horizontal resolutions: 0.01\degree $\times$ 0.01\degree ($\approx$ 1km) and 0.03\degree $\times$ 0.03\degree ($\approx$ 3km). 
 This study only considers the 3km version, as the required reanalysis data are currently available at this resolution. 
 We select CMA-MESO as the baseline model because the long-term objective of this work is to develop a deep learning-based alternative to CMA-MESO for specific tasks.

\subsection{3km Resolution Reanalysis Data}
 The 3km resolution reanalysis data used in this study are produced by China Meteorological Administration Regional Reanalysis Atmospheric System (CMA-RRA). 
 CMA-RRA was developed by the CEMC and is based on the 3km rapid refresh cycling assimilation and forecast system of CMA-MESO.
 Its resolution is 3km/1hour and covers the same region as CMA-MESO.
 It is important to note that this dataset is not yet publicly available.

CMA-RRA comprises multiple modules, including observational data preprocessing and quality control, three-dimensional variational data assimilation, cloud analysis, incremental analysis update (IAU), large-scale background field initialization, multi-scale hybrid assimilation, digital filtering, and mesoscale model forecasting.
CMA-RRA operates through an inner analysis cycle and an outer cycle. 
The inner analysis cycle uses an hourly assimilation-forecast cycle and the hydrometeor variables are not updated during the analysis process.
Using the results from the inner cycle analysis, the outer cycle conducts cloud analysis using networked Doppler weather radar 3D reflectivity products and Fengyun geostationary meteorological satellite cloud products. This process generates cloud analysis and hourly precipitation products.

Evaluated against independent surface station observations, the 3km reanalysis outperforms the CMA-MESO forecasts for near-surface variables. 
For the 2m temperature, it is also markedly superior to ERA5, while the quality of its 10m wind and 2m specific humidity is broadly comparable. 
In the upper air, the 3km reanalysis provides significantly better results than the CMA-MESO forecasts and is generally superior to ERA5 for wind, temperature, and humidity. However, ERA5 performs better in specific cases: the 3km reanalysis exhibits slightly larger errors in mid- to upper-tropospheric (500–100 hPa) winds during spring, slightly larger errors in lower-tropospheric temperature during autumn and winter, and slightly larger errors in humidity near the tropopause during summer.
 
\subsection{Training Setup}
\label{training_setup}
For the training of CorrDiff, we use 25km ERA5 as input data and 3km reanalysis data as target data.
The size of the 25km input data is 192 $\times$ 288 covering the region 12.25-60\degree N, 70-141.75\degree E, and the size of the 3km target data is 1600 $\times$ 2400 covering the region 12.13-60.1\degree N, 70-141.97\degree E.
The mismatch in spatial resolution (25km vs. 3km) (non-divisibility of 25 by 3) results in minor misalignment between input and target region, which could potentially reduce the accuracy on the boundary.
We use the data from 2019 to 2022 as the training set, choosing eight time points each day: 00:00, 03:00, 06:00, 09:00, 12:00, 15:00, 18:00, and 21:00 UTC. In total, the training set contains 11688 samples.

During training, both input and target data are min-max normalized to the interval [-1,1]. The variables on each pressure level are normalized independently, as the distribution of data varies between different pressure levels.
For comparison, we also evaluated normalization via the mean and standard deviation.
The experimental results demonstrate that min-max normalization achieves faster convergence during the early training stage.

For deep learning models, we have the flexibility to select various combinations of variables as inputs and outputs.
In this work, we investigate four distinct input/output configurations, detailed in Table~\ref{input_output}.
For example, for a model that corresponds to Combination 4, the input has 35 variables ($c_{in} = 35$) and the output has 24 variables ($c_{out} = 24$).

\begin{table}[!htbp]
\centering
\caption{Different input/output variable combinations.}
\label{input_output}
\scriptsize
\begin{adjustbox}{width=\textwidth}
\begin{tabular}{|c|c|c|c|c|c|}
\hline
\multirow{2}{*}{Combinations} & 
\multirow{2}{*}{Variables} & 
\multicolumn{2}{c|}{Input} & 
\multicolumn{2}{c|}{Output} \\
\cline{3-6}
& & 
\makecell{Height \\ Levels (m)} & 
Pressure Levels (hPa) & 
\makecell{Height \\ Levels (m)} & 
Pressure Levels (hPa) \\
\hline

\multirow{8}{*}{Combination 1} & 
Zonal Wind (u) & 
10 & 
100 200 500 700 850 925 & 
10 & 
100 200 500 700 850 925 \\
\cline{2-6}
& Meridional Wind (v) & 
10 & 
100 200 500 700 850 925 & 
10 & 
100 200 500 700 850 925 \\
\cline{2-6}
& Geopotential Height (z) & 
- & 
100 200 500 700 850 925 & 
- & 
100 200 500 700 850 925 \\
\cline{2-6}
& Temperature (t) & 
2 & 
100 200 500 700 850 925 & 
2 & 
100 200 500 700 850 925 \\
\cline{2-6}
& Specific Humidity (q) & 
- & 
100 200 500 700 850 925 & 
- & 
100 200 500 700 850 925 \\
\cline{2-6}
& \makecell{Total Column Integrated \\ Water Vapour} & 
- & 
Integrated & 
- & 
Integrated \\
\cline{2-6}
& Mean Sea Level Pressure & 
- & 
Surface & 
- & 
Surface \\
\cline{2-6}
& Surface Pressure & 
- & 
Surface & 
- & 
Surface \\
\hline
\multirow{10}{*}{Combination 2} & 
Zonal Wind (u) & 
10 & 
500 700 850 925 & 
10 & 
500 700 850 925 \\
\cline{2-6}
& Meridional Wind (v) & 
10 & 
500 700 850 925 & 
10 & 
500 700 850 925 \\
\cline{2-6}
& Geopotential Height (z) & 
- & 
500 700 850 925 & 
- & 
500 700 850 925 \\
\cline{2-6}
& Temperature (t) & 
2 & 
500 700 850 925 & 
2 & 
500 700 850 925 \\
\cline{2-6}
& Specific Humidity (q) & 
- & 
500 700 850 925 & 
- & 
500 700 850 925 \\
\cline{2-6}
& \makecell{Total Column Integrated \\ Water Vapour} & 
- & 
Integrated & 
- & 
Integrated \\
\cline{2-6}
& Mean Sea Level Pressure & 
 & 
Surface & 
- & 
Surface \\
\cline{2-6}
& Surface Pressure & 
- & 
Surface & 
- & 
Surface \\
\cline{2-6}
& Orography & 
- & 
Surface & 
- & 
- \\
\cline{2-6}
& Radar Composite Reflectivity & 
- & 
- & 
- & 
Surface \\
\hline

\multirow{9}{*}{Combination 3} & 
Zonal Wind (u) & 
10 & 
100 200 500 700 850 925 & 
10 & 
500 700 850 925 \\
\cline{2-6}
& Meridional Wind (v) & 
10 & 
100 200 500 700 850 925 & 
10 & 
500 700 850 925 \\
\cline{2-6}
& Geopotential Height (z) & 
- & 
100 200 500 700 850 925 & 
- & 
500 700 850 925 \\
\cline{2-6}
& Temperature (t) & 
2 & 
100 200 500 700 850 925 & 
2 & 
500 700 850 925 \\
\cline{2-6}
& Specific Humidity (q) & 
- & 
100 200 500 700 850 925 & 
- & 
500 700 850 925 \\
\cline{2-6}
& \makecell{Total Column Integrated \\ Water Vapour} & 
- & 
Integrated & 
- & 
Integrated \\
\cline{2-6}
& Mean Sea Level Pressure & 
 & 
Surface & 
- & 
Surface \\
\cline{2-6}
& Surface Pressure & 
- & 
Surface & 
- & 
Surface \\
\cline{2-6}
& Orography & 
- & 
Surface & 
- & 
- \\
\hline
\multirow{7}{*}{Combination 4} & 
Zonal Wind (u) & 
10 & 
100 200 500 700 850 925 & 
10 & 
500 700 850 925 \\
\cline{2-6}
& Meridional Wind (v) & 
10 & 
100 200 500 700 850 925 & 
10 & 
500 700 850 925 \\
\cline{2-6}
& Geopotential Height (z) & 
- & 
100 200 500 700 850 925 & 
- & 
500 700 850 925 \\
\cline{2-6}
& Temperature (t) & 
2 & 
100 200 500 700 850 925 & 
2 & 
500 700 850 925 \\
\cline{2-6}
& Specific Humidity (q) & 
- & 
100 200 500 700 850 925 & 
- & 
500 700 850 925 \\
\cline{2-6}
& Mean Sea Level Pressure & 
 & 
Surface & 
- & 
Surface \\
\cline{2-6}
& Orography & 
- & 
Surface & 
- & 
- \\
\hline
\end{tabular}
\end{adjustbox}
\end{table}

\section{Results}
\label{result}
\subsection{Training Results on the Validation Set}
The data from January, April, July, and October 2023 are used as the validation set.
We select eight time points each day as in the training set: 00:00, 03:00, 06:00, 09:00, 12:00, 15:00, 18:00, and 21:00 UTC.
In total, there are 984 samples in the validation set.

\subsubsection{Regression Models}
Five regression models have been trained, denoted as Regression 1, Regression 2, Regression 3-1, Regression 3-2, and Regression 4.
Regression 1, 2, and 4 correspond to Combination 1, 2, and 4 in Table~\ref{input_output}, respectively. Regression 3-1 and 3-2 correspond to Combination 3.
\begin{itemize}
\item Regression 1 is the first trained model. 
It incorporates the most variables to test the feasibility of downscaling multiple variables with a single model.
In order to ensure sufficient model complexity, we increase the size of Regression 1 such that only one sample can be computed at one time on a single NVIDIA H20 GPU with the use of checkpointing.
The UNet embedding size of Regression 1 is [128, 256, 512, 512, 1024].

\item Regression 2 builds upon Regression 1 by adding orography as an additional input variable in order to increase accuracy. 
Furthermore, radar composite reflectivity is added as an output for Regression 2, while high-altitude variables (100 and 200 hPa) are removed for both input and output, since high-altitude forecasting is not our main focus.

\item Regression 3-1 modifies Regression 2 by excluding radar reflectivity as an output but reintroducing the 100 and 200 hPa variables as inputs, under the assumption that additional inputs would not harm performance.
Regression 3-2 is a smaller version of Regression 3-1 with reduced model complexity, designed to test if a reduced-size network could maintain comparable accuracy.
The UNet embedding size of Regression 3-2 is [32, 64, 128, 256, 256].

\item Regression 4 is implemented based on Regression 3-1 but excludes total column integrated water vapour and surface pressure to ensure compatibility with SFF (see Section~\ref{sff}), which does not produce these two variables.
For Regression 1, 2, 3-1 and 3-2, a batch size of 64 is used to ensure stable gradient estimations, while for Regression 4, we experimented with a smaller batch size (a batch size of 8) to explore its effects on training dynamics, acknowledging that this modification would decrease the time per step due to our use of gradient accumulation.
\end{itemize}

The training time per epoch on 8 NVIDIA H20 GPUs is approximately 3.5–4 hours for Regression 1, 2, and 3-1; 1.5 hours for Regression 3-2; and 3 hours for Regression 5.

\begin{figure}[!htbp]
\centering
\includegraphics[width=1\linewidth]{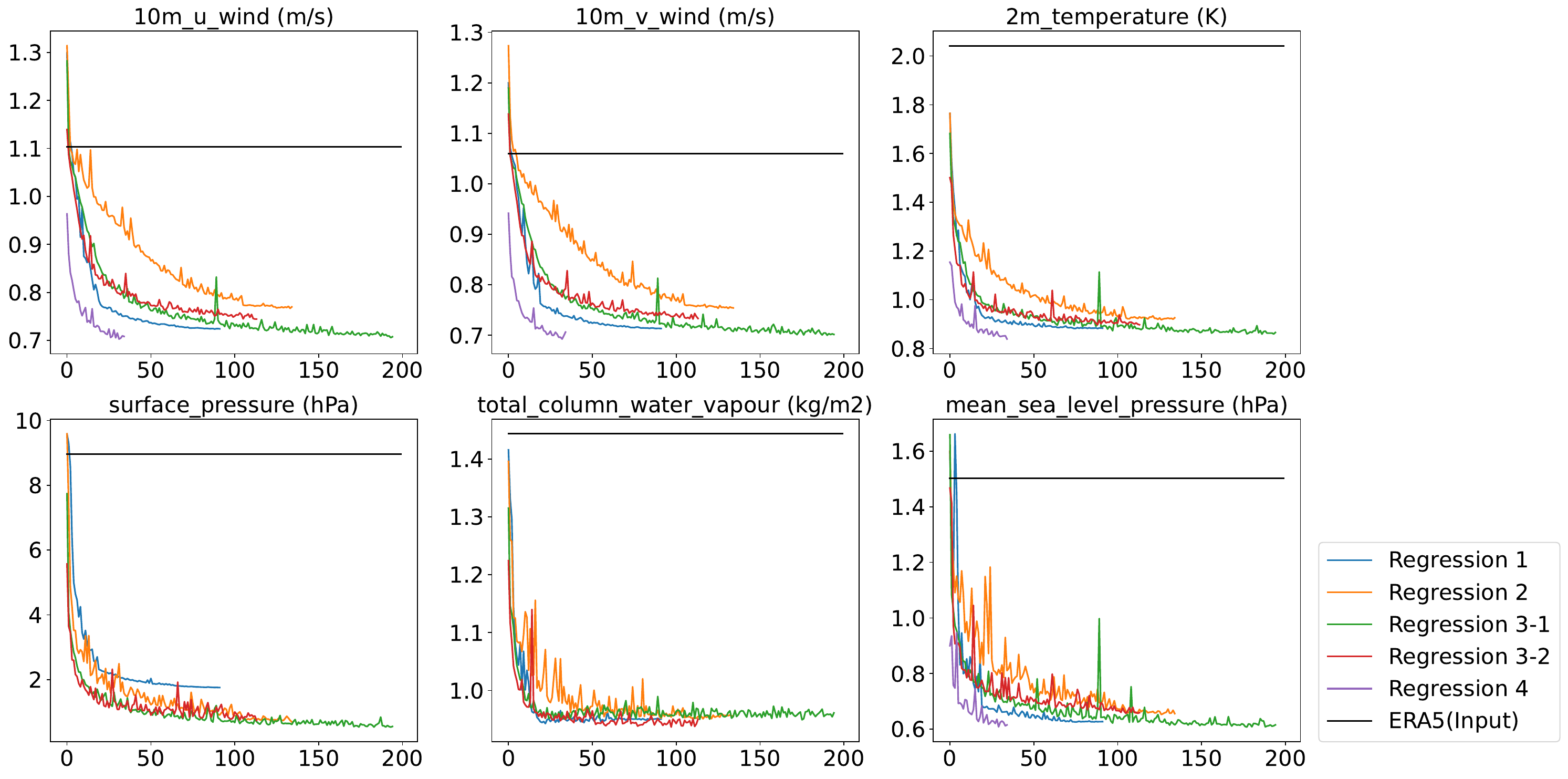}
\caption{Validation curves of the regression models for surface variables. Each subplot corresponds to a target variable. The curves plot the mean absolute error (MAE) (y-axis) against the number of training epochs (x-axis). ERA5 is used as the ground truth.} \label{fig_val_curve_ground}
\end{figure}

\begin{figure}[!htbp]
\centering
\includegraphics[width=1\linewidth]{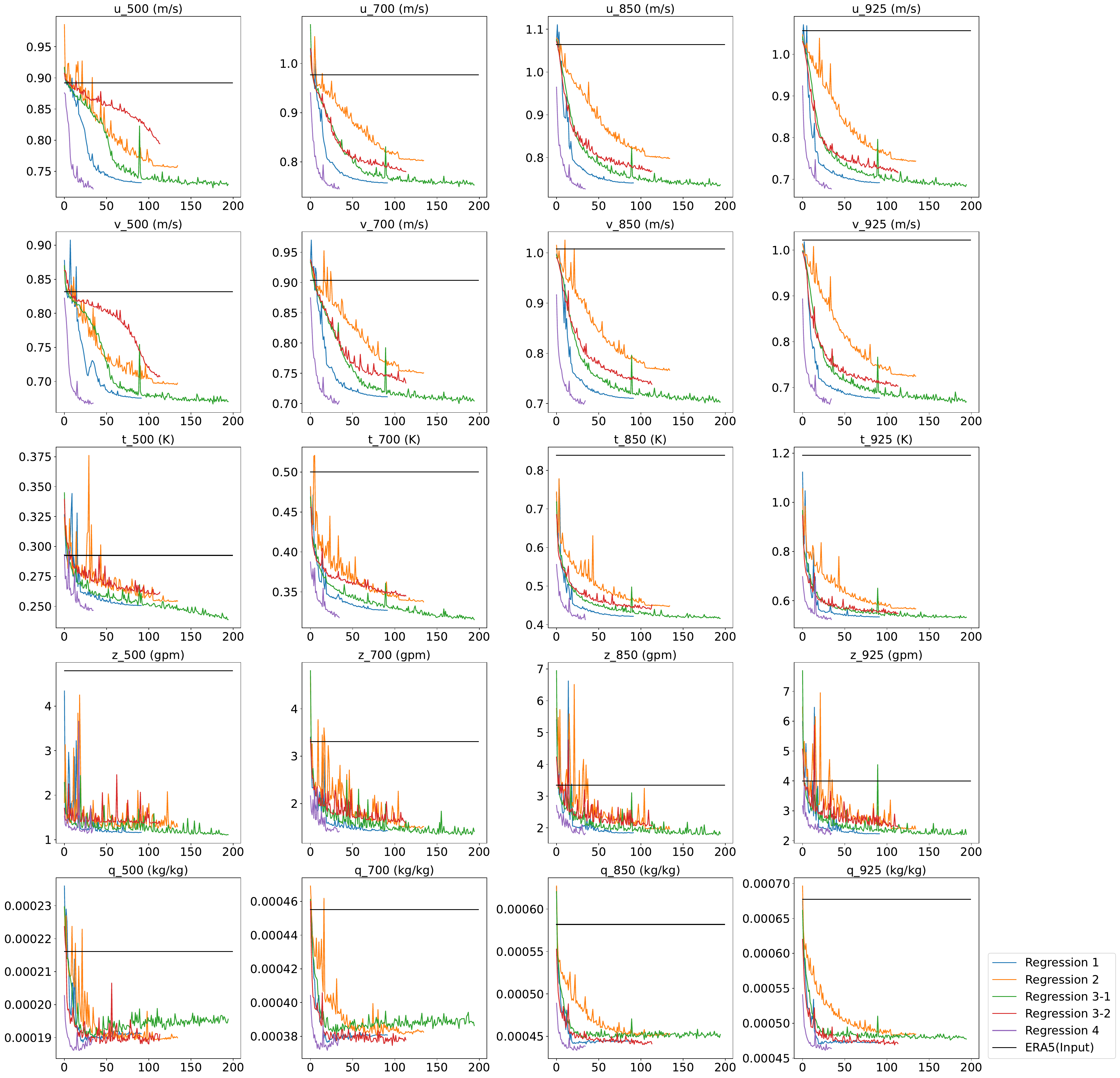}
\caption{Validation curves of regression models for pressure level variables. Each subplot corresponds to a target variable. The curves plot the mean absolute error (MAE) (y-axis) against the number of training epochs (x-axis). ERA5 is used as the ground truth.} \label{fig_val_curve_plevel}
\end{figure}

The validation curves for the five regression models are shown in Fig. \ref{fig_val_curve_ground} (for surface variables) and Fig. \ref{fig_val_curve_plevel} (for pressure level variables).
Since only Regression 2 predicts radar reflectivity, the validation curve of radar reflectivity is excluded due to the absence of comparative benchmarks.

In each subplot, the black horizontal line denotes the MAE of the ERA5 input for a variable (MAE between $bilinear(x_i^u)$ and $y_i^v$ for all ($x_i$, $y_i$) in the validation set, where $x_i^u$ and $y_i^v$ correspond to this variable). 
After the first few epochs, all validation curves remain below the black line, indicating that all regression models consistently outperform simple interpolation.
The validation curves exhibit distinctive patterns between variables.
For example, the validation curves of the 10m wind decrease with a progressively decreasing rate of decline, while, for the 500 hPa wind, the rate of decline decreases at first and increases after several epochs.
In addition, all validation curves display a globally decreasing trend, except for specific humidity (q), suggesting the existence of overfitting for this variable. 
These inter-variable differences imply that employing variable-specific embedding methods could potentially enhance both the convergence speed and model accuracy.

The comparative analysis between the validation curves of Regression 1 and Regression 2 shows that Regression 1 significantly outperforms Regression 2 on all variables except surface pressure. 
It is important to note that we did not exhaustively explore all combinations, for example, between Combination 1 and Combination 2, we can also examine another combination that does not include radar composite reflectivity as well as the 100 and 200 hPa variables.
We assume that the impact of the 100 and 200 hPa variables on the downscaling of other variables exists but is minor and that including orography as an input should not harm the performance.
Therefore, the results suggest that integrating radar reflectivity inference with downscaling in a single model might compromise downscaling accuracy, and adding orography to the inputs increases the accuracy of surface pressure.

As expected, Regression 3-1, which excludes radar composite reflectivity but includes orography as an additional input, outperforms both Regression 1 and 2. 
The reduction of model complexity in Regression 3-2 causes a performance degradation for most variables, indicating that this reduced version does not have sufficient model complexity.

The curves of Regression 4 are significantly lower than those of the previous models.
This improvement can be attributed to two potential reasons: first, the smaller batch size may provide better gradient estimations for this downscaling task; 
second, excluding surface pressure and total column integrated water vapour likely reduces task complexity, freeing up model capacity for the remaining variables.
Consequently, further adjustment to the batch size and a more sophisticated selection of input/output variables could lead to additional performance gains.

In the rest of this paper, we mainly focus on analyzing Regression 2 and Regression 4, as only Regression 2 can output radar composite reflectivity, and Regression 4 is compatible with SFF and has the highest accuracy on the validation set.

\begin{figure}[!htbp]
\centering
\includegraphics[width=1\linewidth]{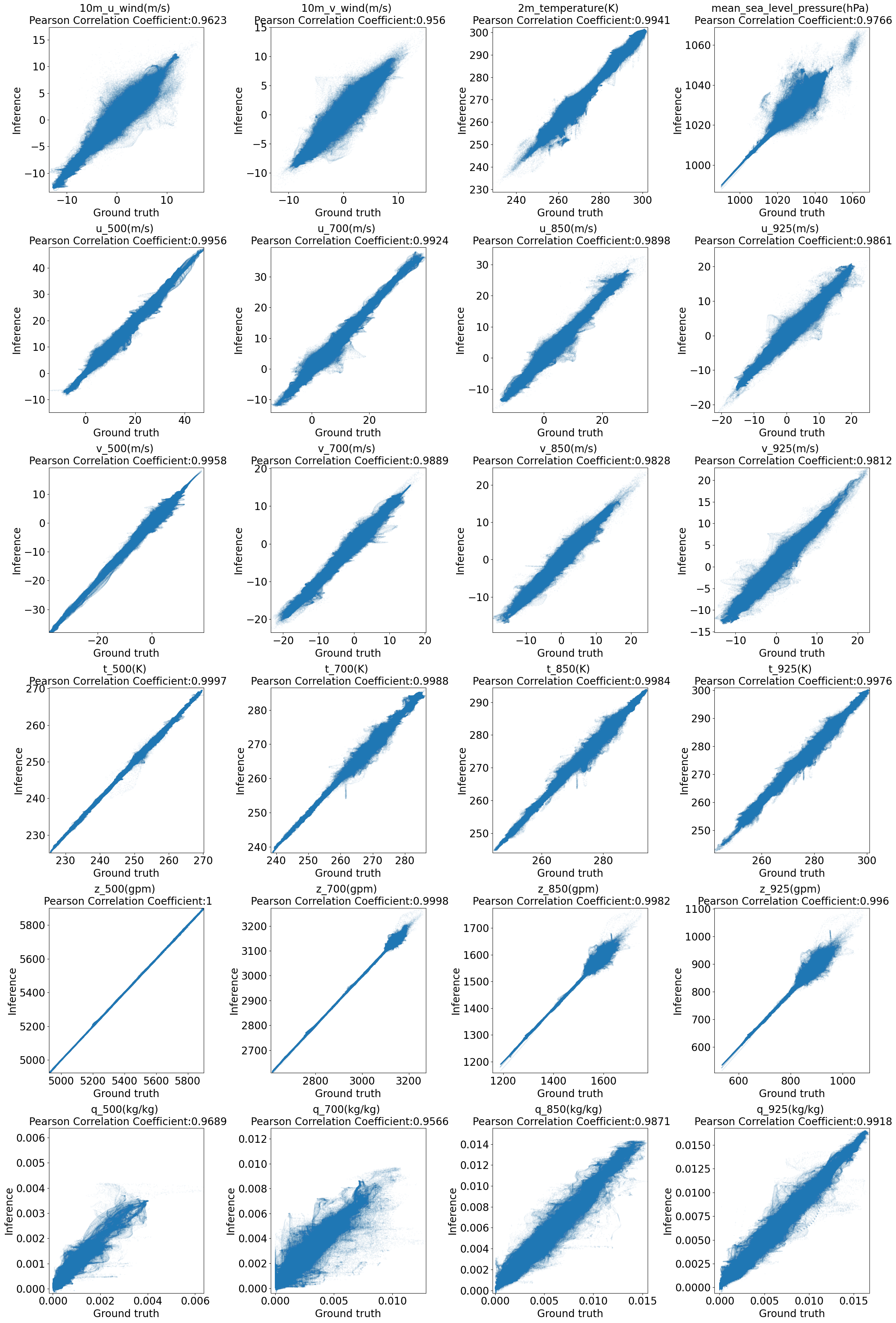}
\caption{Correlation between ground truth and the inference of Regression 4 for data from 2023-03-01-00 UTC.} \label{fig_regression_scatter}
\end{figure}

Fig. \ref{fig_regression_scatter} shows the correlation between the ground truth (ERA5) and the predictions of Regression 4 for data from 2023-03-01-00 UTC.
For wind, temperature, and geopotential height, the Pearson correlation coefficients for the high-level variables are consistently higher than those of the surface variables.
In fact, high-level variables are typically dominated by smooth large-scale weather patterns, while low-level variables are influenced by local and complex interactions with the Earth's surface, which consequently have more high-frequency patterns.
So, downscaling high-level variables might be easier than downscaling surface and low-level variables for statistical models.
These findings suggest that model performance could potentially be improved by not treating all variables equally. 
For instance, increasing model capacity for surface variables (for example, through a specialized network structure) or employing a weighted loss function might lead to performance gains. 

\subsubsection{CorrDiff Models}
The CorrDiff models that correspond to Regression 1, 2, and 4 are denoted as CorrDiff 1, 2, and 4, respectively.
Their MAE scores on the validation set are presented in Table~\ref{MAE_val}, where Regression 1, 2, and 4 were trained for 91, 120 and 35 epochs, respectively, and CorrDiff 1, 2, and 4 were trained for 285, 100 and 59 epochs, respectively.
All reported MAE scores for the CorrDiff models are computed using a single sample.
Our results indicate that the MAE scores of the regression models are consistently lower than those of the CorrDiff models.
In fact, the degradation in MAE of the CorrDiff models compared to that of the regression models is also reported in \cite{mardani2025residual}.
As stated in \cite{mardani2025residual}, the degradation is expected as the diffusion models optimize the Kullback-Leibler divergence as opposed to the regression models that minimize the MSE loss.
Following \cite{mardani2025residual}, we also compute the continuous ranked probability score (CRPS) \cite{hersbach2000decomposition}, see Table~\ref{MAE_crps}.
CRPS can be considered as a generalization of the mean absolute error for probability prediction.
Since the computation of CRPS is slow with our current implementation (primarily due to the large grid size), we select a more restricted set of time points for its calculation: 00:00 and 12:00 UTC on the first day of each month in 2023 and we only focus on CorrDiff 4. As expected, the CRPS values are generally the lowest.

\begin{figure}[!htbp]
\centering
\includegraphics[width=1\linewidth]{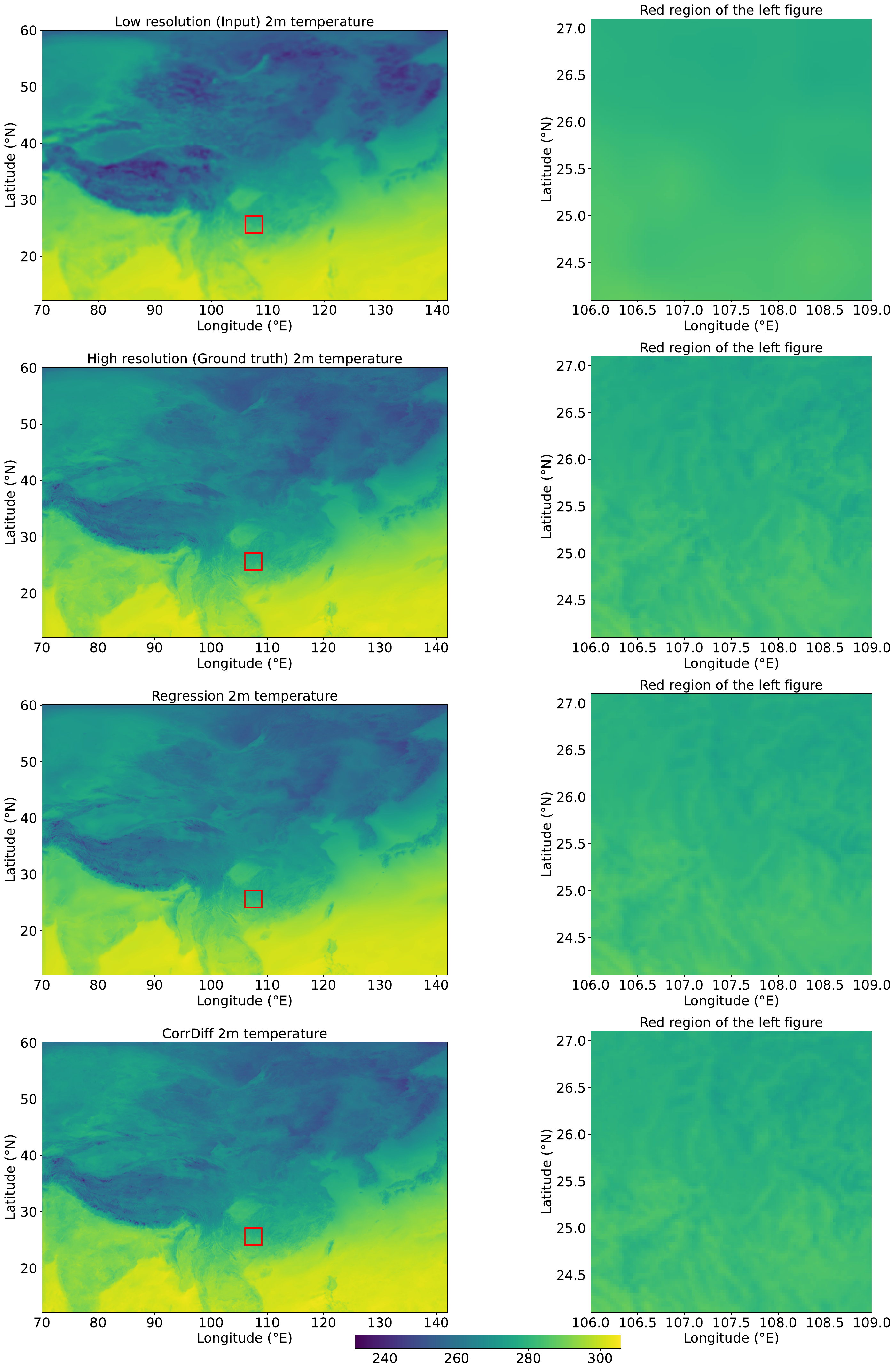}
\caption{Illustration of the 2m temperature inference of Regression 4 and CorrDiff 4 for data from 2023-12-01-00 UTC. Each figure on the right is a zoomed view of the red box area in the left figure.} \label{fig_qualitative_2mt}
\end{figure}

\begin{figure}[!htbp]
\centering
\includegraphics[width=1\linewidth]{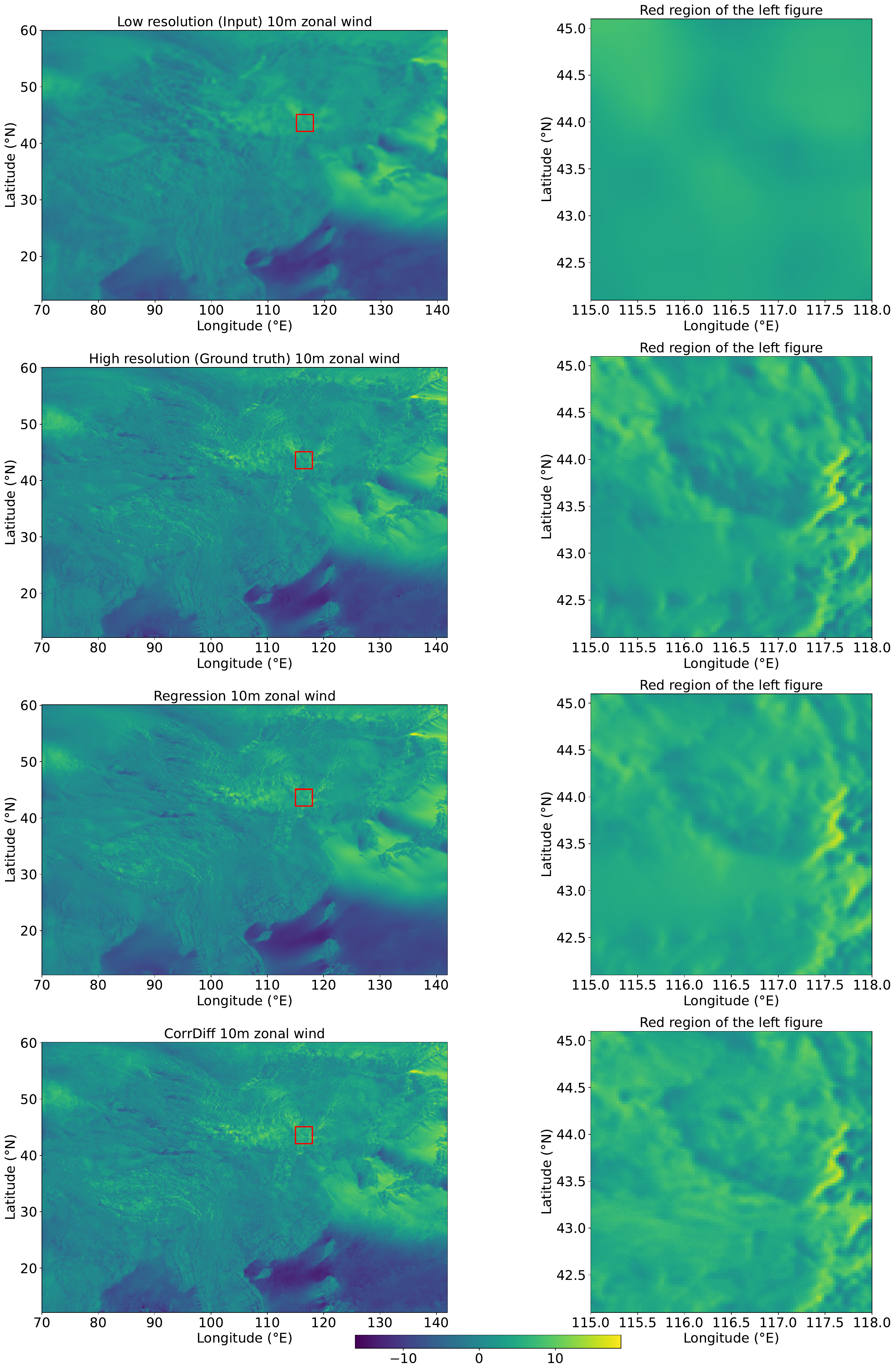}
\caption{Illustration of the 10m zonal wind inference of Regression 4 and CorrDiff 4 for data from 2023-12-01-00 UTC. Each figure on the right is a zoomed view of the red box area in the left figure.} \label{fig_qualitative_10mu}
\end{figure}

\begin{figure}[!htbp]
\centering
\includegraphics[width=0.7\linewidth]{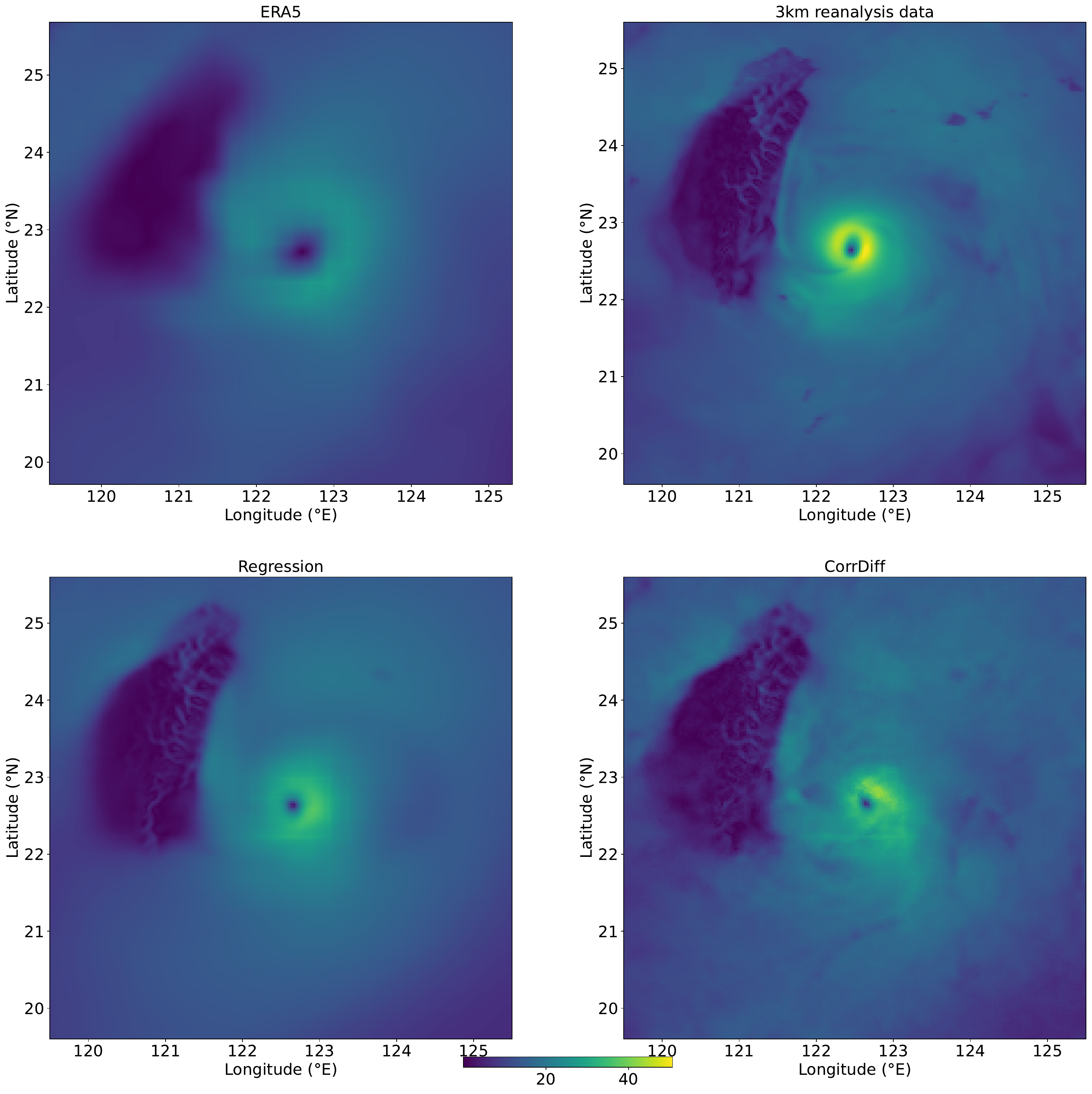}
\caption{Illustration of the downscaling of Typhoon Haikui on 2023-09-03-00 UTC by Regression 4 and CorrDiff 4. Figures show the 10m wind speed.} \label{fig_haikui}
\end{figure}

Some examples of the predictions of Regression 4 and CorrDiff 4 are given in Fig. \ref{fig_qualitative_2mt} and Fig. \ref{fig_qualitative_10mu}.
The local patterns that do not occur in the low-resolution data (top-right) are captured by both the regression and CorrDiff models.
An example of typhoon downscaling is given in Fig. \ref{fig_haikui}. The typhoon structures generated by both Regression 4 and CorrDiff 4 are more similar to the 3km reanalysis data compared to ERA5.
The comparison between the outputs of Regression 4 and CorrDiff 4 reveals that the CorrDiff models generate more high-frequency details.




\begin{table}[!htbp]
\centering
\caption{MAE on the validation set.}
\begin{adjustbox}{width=\textwidth}
\label{MAE_val}
\begin{tabular}{|c|c|c|c|c|c|c|c|}
\hline
Variable (Unit) & ERA5 (Input) & Regression 1 & Regression 2 & Regression 4 & CorrDiff 1 & CorrDiff 2 & CorrDiff 4 \\
\hline
\makecell{10m Zonal \\ Wind (m/s)}        & 1.1    & 0.72   & 0.77   & \textbf{0.7}    & 1.05 & 1.25 & 0.98   \\
\hline
\makecell{10m Meridional \\ Wind (m/s)}   & 1.05   & 0.71  & 0.75  & \textbf{0.70}    & 1.04 & 1.16 & 0.97   \\
\hline
2m Temperature (K)           & 2.04   & 0.88    & 0.92    & \textbf{0.83}   & 1.10 & 1.21 & 1.05   \\
\hline
\makecell{Mean Sea \\ Level Pressure (hPa)}    & 1.5    & 0.62  & 0.66   & \textbf{0.61}   & 0.88 & 1.07 & 0.81   \\
\hline
Surface Pressure (hPa)          & 8.96   & 1.75   & \textbf{0.77}   & {--}   & 1.88 & 2.01 & {--}   \\
\hline
\makecell{Total Column \\ Integrated \\ Water Vapour (kg/m2)} & 1.44   & \textbf{0.95}   & 0.95   & {--}   & 1.29 & 1.47 & {--}   \\
\hline
u100 (m/s)                      & 0.79   & \textbf{0.71}  & {--}        & {--}   & 1.10 & {--}        & {--}   \\
\hline
u200 (m/s)                      & 0.99   & \textbf{0.94}   & {--}        & {--}   & 1.43 & {--}        & {--}   \\
\hline
u500 (m/s)                      & 0.89   & 0.73  & 0.75   & \textbf{0.72}   & 1.11 & 1.21 & 1.06   \\
\hline
u700 (m/s)                      & 0.97   & 0.75   & 0.80   & \textbf{0.74}   & 1.13 & 1.31  & 1.08   \\
\hline
u850 (m/s)                      & 1.06   & 0.74   & 0.80  & \textbf{0.72}   & 1.09 & 1.28 & 1.03   \\
\hline
u925 (m/s)                      & 1.05   & 0.69   & 0.74   & \textbf{0.67}   & 1.01 & 1.13 & 0.98   \\
\hline
v100 (m/s)                      & 0.78   & \textbf{0.63}   & {--}        & {--}   & 0.95 & {--}        & {--}   \\
\hline
v200 (m/s)                      & 1.00      & \textbf{0.89}   & {--}        & {--}   & 1.35  & {--}        & {--}   \\
\hline
v500 (m/s)                      & 0.83   & 0.67  & 0.69  & \textbf{0.66}   & 1.02 & 1.08 & 0.96   \\
\hline
v700 (m/s)                      & 0.90    & 0.71   & 0.75   & \textbf{0.70}    & 1.08 & 1.22 & 1.02   \\
\hline
v850 (m/s)                      & 1.00      & 0.71   & 0.76   & \textbf{0.70}    & 1.07 & 1.25 & 1.01   \\
\hline
v925 (m/s)                      & 1.02   & 0.67  & 0.72  & \textbf{0.67}   & 0.99 & 1.14 & 0.96   \\
\hline
z100 (gpm)                      & 12.91  & \textbf{2.08}   & {--}        & {--}   & 4.91 & {--}        & {--}   \\
\hline
z200 (gpm)                      & 9.49   & \textbf{1.75}   & {--}        & {--}   & 4.42 & {--}        & {--}   \\
\hline
z500 (gpm)                      & 4.79   & \textbf{1.16}   & 1.53   & 1.27   & 2.67 & 3.44 & 2.55   \\
\hline
z700 (gpm)                      & 3.30    & 1.43   & 1.50   & \textbf{1.41}   & 2.74 & 3.23  & 2.43   \\
\hline
z850 (gpm)                      & 3.34   & 1.84   & 1.96   & \textbf{1.82}   & 3.45 & 3.97 & 3.28   \\
\hline
z925 (gpm)                      & 3.99   & \textbf{2.23}   & 2.39   & 2.24   & 3.97 & 6.49 & 4.33   \\
\hline
t100 (K)                      & 0.58   & \textbf{0.45}   & {--}        & {--}   & 0.64 & {--}        & {--}   \\
\hline
t200 (K)                      & 0.36   & \textbf{0.29}  & {--}        & {--}   & 0.44 & {--}        & {--}   \\
\hline
t500 (K)                      & 0.29   & 0.25   & 0.25  & \textbf{0.24}   & 0.40 & 0.46 & 0.38   \\
\hline
t700 (K)                      & 0.5    & 0.32   & 0.33   & \textbf{0.31}   & 0.49 & 0.56  & 0.45   \\
\hline
t850 (K)                      & 0.83   & 0.42   & 0.45  & \textbf{0.41}   & 0.61 & 0.68 & 0.57   \\
\hline
t925 (K)                      & 1.19   & 0.53  & 0.56   & \textbf{0.52}   & 0.75 & 0.92 & 0.69   \\
\hline
q100 (kg/kg)                      & $8.55 \times 10^{-7}$ & \bm{$6.90\times10^{-7}$}     & {--}        & {--}   & $1.32\times 10^{-6}$  & {--}        & {--}   \\
\hline
q200 (kg/kg)                      & $7.52 \times 10^{-6}$ & \bm{$6.86\times 10^{-6}$}     & {--}        & {--}   & $1.06 \times 10^{-5}$  & {--}        & {--}   \\
\hline
q500 (kg/kg)                      & 0.00021 & 0.00019 & 0.00018 & \textbf{0.00018} & 0.00026 & 0.00028 & 0.00025 \\
\hline
q700 (kg/kg)                      & 0.00045 & 0.00038 & 0.00038 & \textbf{0.00037} & 0.00051 & 0.00058 & 0.00052 \\
\hline
q850 (kg/kg)                      & 0.00058 & 0.00044 & 0.00045 & \textbf{0.00043} & 0.00060 & 0.00068  & 0.00058 \\
\hline
q925 (kg/kg)                      & 0.00067 & 0.00047 & 0.00048  & \textbf{0.00046} & 0.00062 & 0.00071 & 0.00061 \\
\hline
\makecell{Radar Composite \\ Reflectivity (dBz)} & {--}   & {--}        & 6.64   & {--}   & {--}        & 8.17 & {--}   \\
\hline
\end{tabular}
\end{adjustbox}
\end{table}

\begin{table}[!htbp]
\centering
\caption{MAE and CRPS for the data at 00:00 and 12:00 UTC on the first day of each month in 2023 of Regression 4 and CorrDiff 4.}
\label{MAE_crps}
\begin{adjustbox}{width=\textwidth}
\begin{tabular}{|c|c|c|c|}
\hline
Variable (Unit) & Regression 4 (MAE) & CorrDiff 4 (MAE) & CorrDiff 4 (CRPS) \\
\hline
10m Zonal Wind (m/s) & 0.73 & 1.02 & \textbf{0.55} \\ 
\hline
10m Meridional Wind (m/s) & 0.73 & 1.01 & \textbf{0.55} \\ \hline
2m Temperature (K) & 0.88 & 1.10 & \textbf{0.66} \\ \hline
Mean Sea Level Pressure (hPa) & 0.58 & 0.78 & \textbf{0.45} \\ \hline
u500 (m/s) & 0.73 & 1.09 & \textbf{0.56} \\ \hline
u700 (m/s) & 0.76 & 1.10 & \textbf{0.59} \\ \hline
u850 (m/s) & 0.75 & 1.08 & \textbf{0.57} \\ \hline
u925 (m/s) & 0.70 & 1.01 & \textbf{0.53} \\ \hline
v500 (m/s) & 0.68 & 0.99 & \textbf{0.52} \\ \hline
v700 (m/s) & 0.72 & 1.06 & \textbf{0.55} \\ \hline
v850 (m/s) & 0.73 & 1.05 & \textbf{0.55} \\ \hline
v925 (m/s) & 0.70 & 1.02 & \textbf{0.54} \\ \hline
t500 (K) & 0.25 & 0.38 & \textbf{0.20} \\ \hline
t700 (K) & 0.32 & 0.46 & \textbf{0.24} \\ \hline
t850 (K) & 0.41 & 0.57 & \textbf{0.31} \\ \hline
t925 (K) & 0.52 & 0.70 & \textbf{0.40} \\ \hline
z500 (gpm) & 1.26 & 2.53 & \textbf{1.14} \\ \hline
z700 (gpm) & 1.47 & 2.44 & \textbf{1.22} \\ \hline
z850 (gpm) & 1.84 & 3.33 & \textbf{1.83} \\ \hline
z925 (gpm) & \textbf{2.23} & 4.26 & 2.57 \\ \hline
q500 (kg/kg) & 0.00019 & 0.00026 & \textbf{0.00015} \\ \hline
q700 (kg/kg) & 0.00039 & 0.00054 & \textbf{0.00031} \\ \hline
q850 (kg/kg) & 0.00045 & 0.00060 & \textbf{0.00035} \\ \hline
q925 (kg/kg) & 0.00048 & 0.00064 & \textbf{0.00038} \\ \hline

\end{tabular}
\end{adjustbox}
\end{table}

\subsubsection{Uncertainty Estimation based on CorrDiff}
\begin{figure}[!htbp]
\centering
\includegraphics[width=1\linewidth]{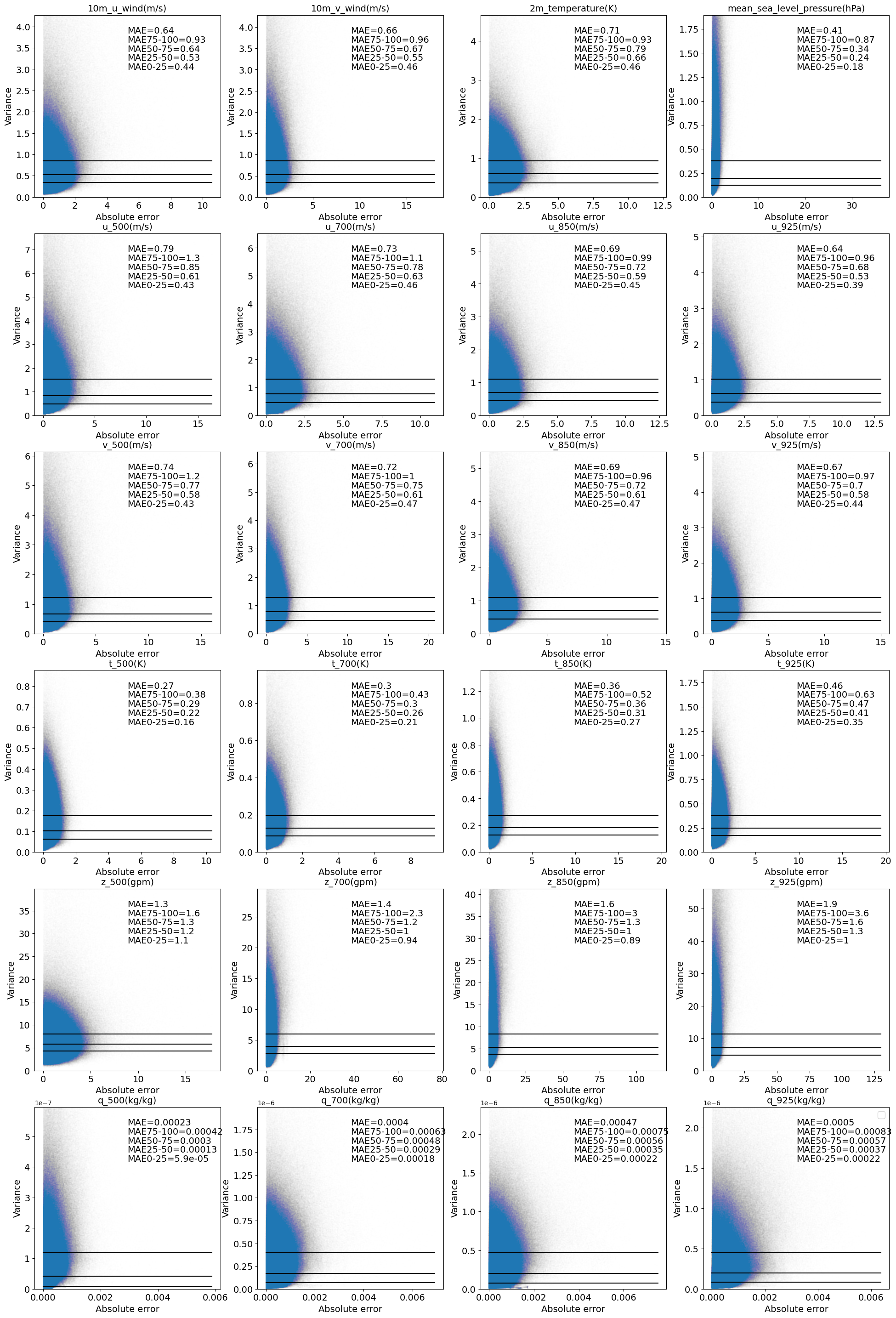}
\caption{Correlation between the absolute error of Regression 4 and the variance of 20-member CorrDiff 4 predictions for the data from 2023-10-01-00 UTC. 
Each point in the figure corresponds to a grid point.
The three black lines in each subplot represent the 75th, 50th and 25th percentile of the variance. These three lines separate the grid points into four groups. MAE75-100 is the MAE of the grid points above the 75th line, MAE50-75 is the MAE of the grid points between the 50th and 75th line, etc.} \label{fig_uncertainty_scatter}
\end{figure}

A key advantage of diffusion models is that they can provide probabilistic predictions.
The variance of the N-member ensemble predictions from CorrDiff can be used to quantify the predictive uncertainty.
Fig. \ref{fig_uncertainty_scatter} illustrates the correlation between the absolute errors of Regression 4 and the variance of the 20-member CorrDiff 4 predictions for the data from 2023-10-01-00 UTC.
We study Regression 4 as it achieves the best MAE scores.

In each subplot, the x- and y-axis show the absolute error and variance at each grid point, respectively.
The shape of the point sets reveals that grid points that exhibit both high absolute error and very low variance are rare, suggesting that low variance is a potential indicator of high accuracy.
Due to the high density of points, it is difficult to clearly see the distribution of the points. So we plot three horizontal black lines that mark the 75th, 50th, and 25th percentile of the variance.
These percentiles partition the data into four groups, each representing a different level of uncertainty (for example, points above the 75th percentile have the highest uncertainty).
The MAE for each group is computed and annotated within the subplots. 
For all variables, the MAE decreases markedly as the variance decreases across these four groups. 
The results demonstrate that variance thresholds can be used to identify regions with higher or lower MAE.

\begin{figure}[!htbp]
\centering
\includegraphics[width=1\linewidth]{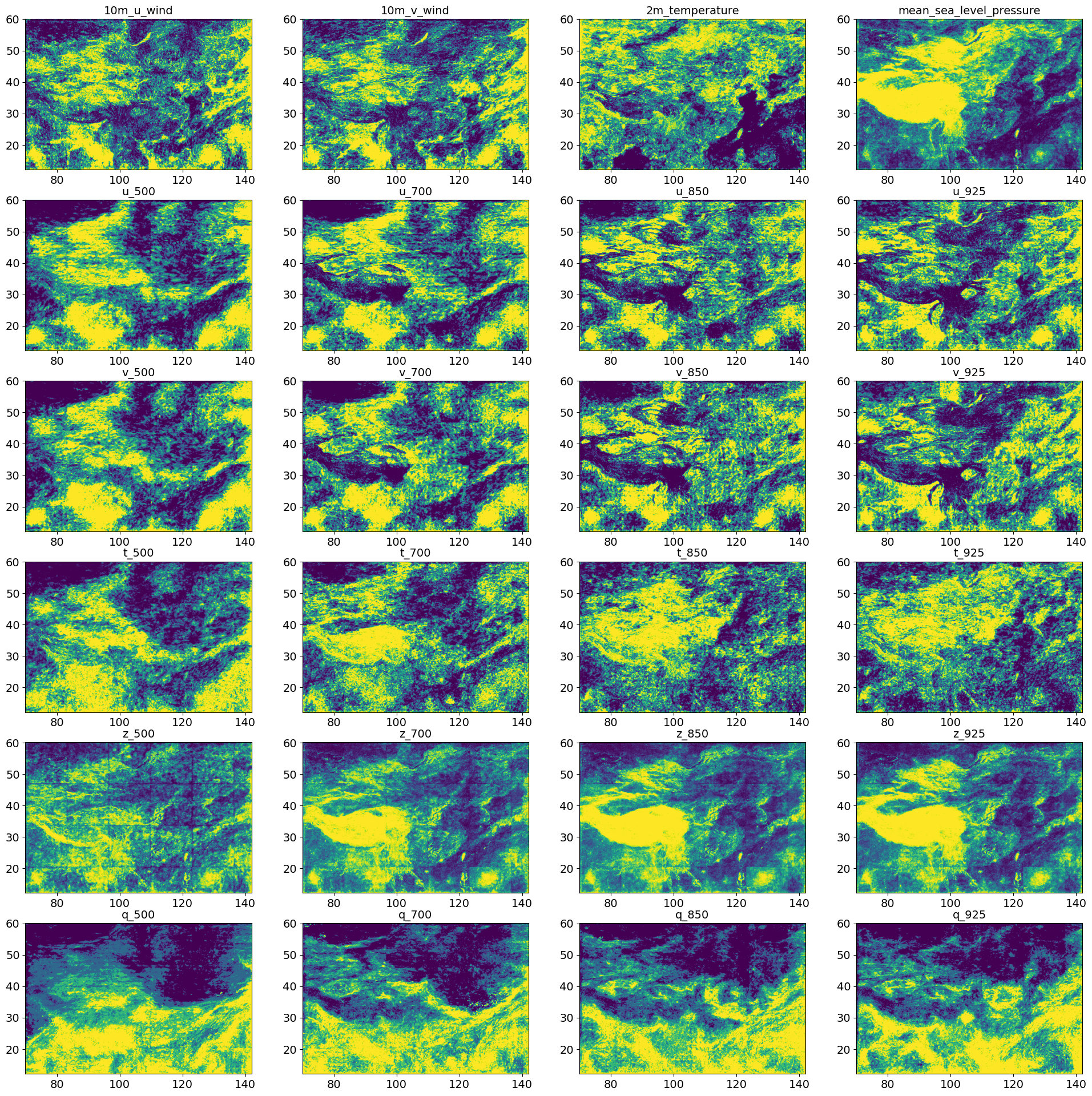}
\caption{Illustration of the four uncertainty groups of grid points in Fig. \ref{fig_uncertainty_scatter}.
Brighter areas correspond to higher variance.} \label{fig_uncertainty_map}
\end{figure}

The spatial distribution of these four uncertainty groups is illustrated in Fig. \ref{fig_uncertainty_map}, where brighter areas represent higher variance.
These results provide a more intuitive understanding of the predictive uncertainty.
For example, the variance of the 2m temperature over the ocean is significantly lower than that over the land; the variance of the geopotential height over the Qinghai-Tibet Plateau is generally higher than that in other regions; in the bottom row of Fig. \ref{fig_uncertainty_map}, the variance in the lower part (which corresponds to areas with more water vapour) of each subplot is higher. 

\subsection{Downscaling the SFF and CMA-GFS Forecasts}
\label{downscaling_gfs}
This section applies the trained downscaling models to the low-resolution outputs of the global forecast systems to generate 3km forecasts.
Two global forecast systems are considered: SFF, a deep learning model, and CMA-GFS, a traditional numerical weather forecast system.
The resulting 3km forecasts are evaluated against the CMA-MESO model, which serves as our baseline model.
The 3km reanalysis data are used as the ground truth.

\subsubsection{SFF}
\label{sff}
Sphere Fusion Forecast (SFF) is a data-driven deep learning-based weather model developed from Spherical Fourier Neural Operators (SFNO) \cite{bonev2023spherical}.
Compared to SFNO, two major improvements are made in SFF: the up-sampling and down-sampling operators between the SFNO blocks are added, allowing the initial and final stages of the SFNO block chain to handle broader frequency spectra, while the middle layers focus on relatively low-frequency information; a Vision Transformer–like architecture between the encoder and decoder is introduced as the skip connection, which improves the model’s capacity for local feature learning, producing more robust and accurate forecasts.

\subsubsection{CMA-GFS}
 China Meteorological Administration-Global Forecast System (CMA-GFS) is a global model developed by the CEMC.
It comprises a semi-implicit semi-Lagrangian (SISL) non-hydrostatic dynamical core, a physical parameterization package, and a four-dimensional variational data assimilation system \cite{chen2008new}. Currently, the horizontal resolution of CMA-GFS is 0.125\degree $\times$ 0.125\degree ($\approx$ 12.5km).

\subsubsection{3km Forecast Evaluation}
\begin{figure}[!htbp]
\centering
\includegraphics[width=1\linewidth]{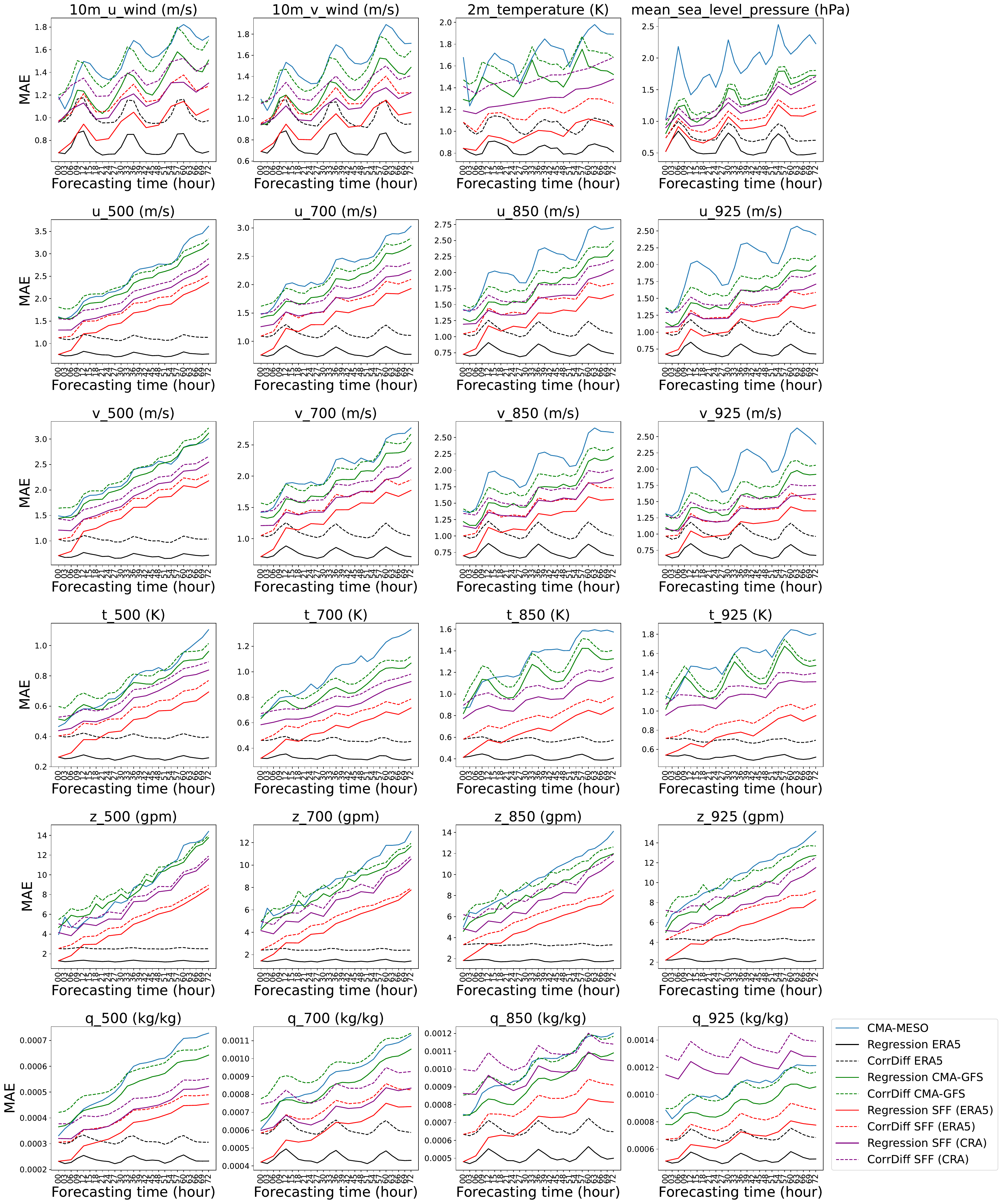}
\caption{The MAE scores of the 3km forecasts that include the downscaled SFF forecasts (taking CRA1.5 or ERA5 as initial fields), the downscaled CMA-GFS forecasts and the CMA-MESO forecasts, using the 3km reanalysis data as the ground truth. 
The downscaled ERA5 are also present for comparison.
The downscaling models are Regression 4 and CorrDiff 4. Seven initialization times are considered: 2023-03-01-00, 2023-06-01-00, 2023-09-01-00, 2023-12-01-00, 2023-05-24-00, 2023-07-30-00, and 2023-09-04-00 UTC.} \label{fig_curve_72}
\end{figure}

Fig. \ref{fig_curve_72} shows the MAE scores for the 3km forecasts that are based on the 25km forecasts of different global models.
Seven initialization times, which are not included in the training set of the CorrDiff models or SFF, are considered: 2023-03-01-00, 2023-06-01-00, 2023-09-01-00, 2023-12-01-00, 2023-05-24-00, 2023-07-30-00, and 2023-09-04-00 UTC. 
The last three times correspond to the periods of Typhoon Mawar, Khanun and Haikui, respectively.
The current operational CMA-MESO provides forecasts up to 72 hours, so we also consider 72 hour forecasts for comparision.
The downscaling results for ERA5 are also given in each subplot, which can be considered as an upper-bound for the downscaling task.
For SFF, forecasts are generated from two initial fields: ERA5 and CRA1.5.
In fact, the current version of SFF is trained on CRA1.5, which is a reanalysis dataset developed by the CMA and is utilized in the AIM-FDP project. 
This project aims to facilitate the deployment of AI models for meteorological forecasting.
The daily updates of CRA1.5 are available 2 hours behind real time. So, the SFF forecasts using CRA1.5 as initial fields can be considered as real-time forecasts.

In each subplot, regression models exhibit lower MAE scores than the corresponding CorrDiff models, which is consistent with the results on the validation set.
In all figures, the curves of the downscaled ERA5 (black curves) exhibit the lowest MAE scores, followed by the downscaled SFF forecasts that take ERA5 as initial fields (red curves).
For most variables except specific humidity, generally, the downscaled SFF forecasts that uses CRA1.5 as initial fields (purple curves) are better than the downscaled CMA-GFS forecasts (green curves), which in turn are better than the CMA-MESO forecasts (blue curves). 
In general, these results in Fig. \ref{fig_curve_72} indicate that, in terms of the MAE scores, our data-driven models can potentially outperform CMA-MESO for most variables.


\begin{figure}[!htbp]
\centering
\includegraphics[width=1\linewidth]{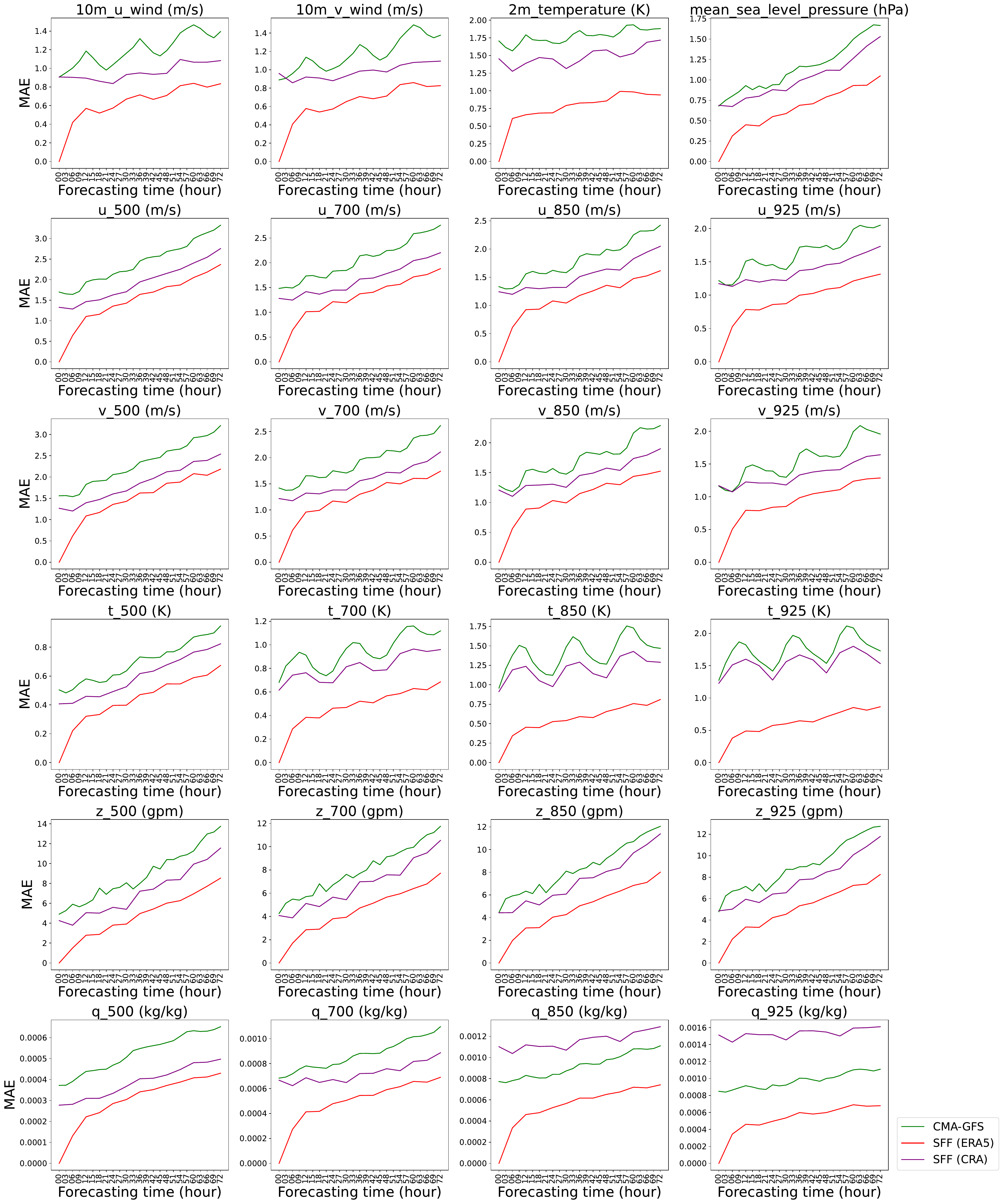}
\caption{MAE between ERA5 and the 25km forecasts of SFF (taking CRA1.5 or ERA5 as initial fields) and CMA-GFS. Seven initialization times are considered: 2023-03-01-00, 2023-06-01-00, 2023-09-01-00, 2023-12-01-00, 2023-05-24-00, 2023-07-30-00, and 2023-09-04-00 UTC.} \label{fig_curve_72_25km}
\end{figure}

The performance of downscaling models is highly dependent on the quality of the input low-resolution data.
Fig. \ref{fig_curve_72_25km} shows the MAE scores of the 25km forecasts of CMA-GFS and SFF, using ERA5 as the ground truth.  
The relationship between the curves of the downscaled CMA-GFS forecasts, downscaled SFF forecasts that use ERA5 as initial fields, and dowscaled SFF forecasts that use CRA1.5 as initial fields in Fig. \ref{fig_curve_72} is consistent with the relationship between the curves of the CMA-GFS forecasts, SFF forecasts that use ERA5 as initial fields, and SFF forecasts that use CRA1.5 as initial fields in Fig. \ref{fig_curve_72_25km} for each subplot.
These results indicate that the errors in the outputs of the downscaling models are mainly determined by the errors present in the low-resolution inputs.

\begin{figure}[!htbp]
\centering
\includegraphics[width=0.8\linewidth]{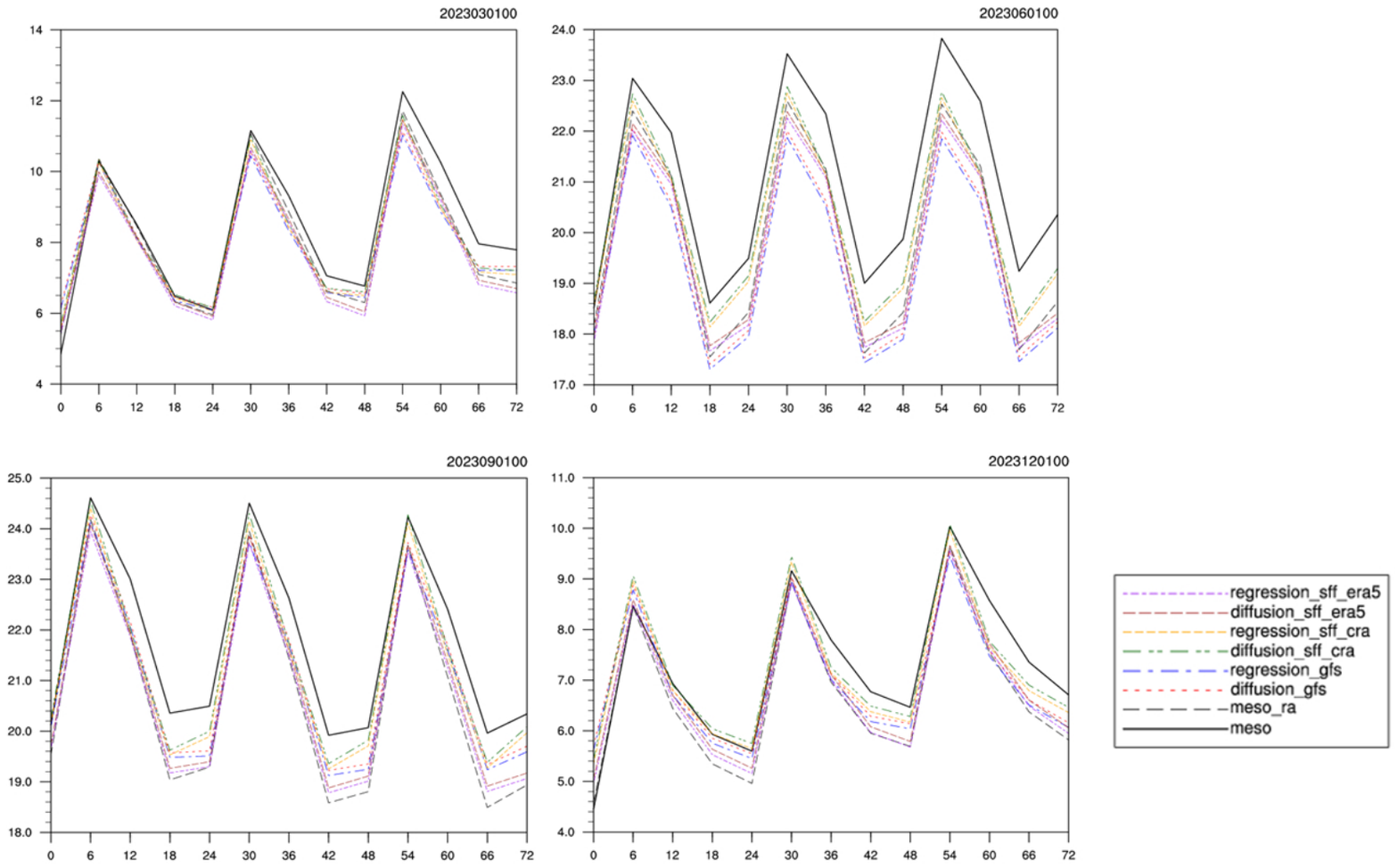}
\caption{Area average of the 2m temperature of the 3km reanalysis data and the 3km forecasts of CMA-MESO, downscaled SFF (taking CRA1.5 or ERA5 as initial fields), and downscaled CMA-GFS. Four initialization times are considered: 2023-03-01-00, 2023-06-01-00, 2023-09-01-00, and 2023-12-01-00 UTC. "meso\_ra" denotes the 3km reanalysis; "regression" and "diffusion" refer to the Regression 4 and CorrDiff 4, respectively.
} \label{fig_area_ave_2mt}
\end{figure}

\begin{figure}[!htbp]
\centering
\includegraphics[width=0.8\linewidth]{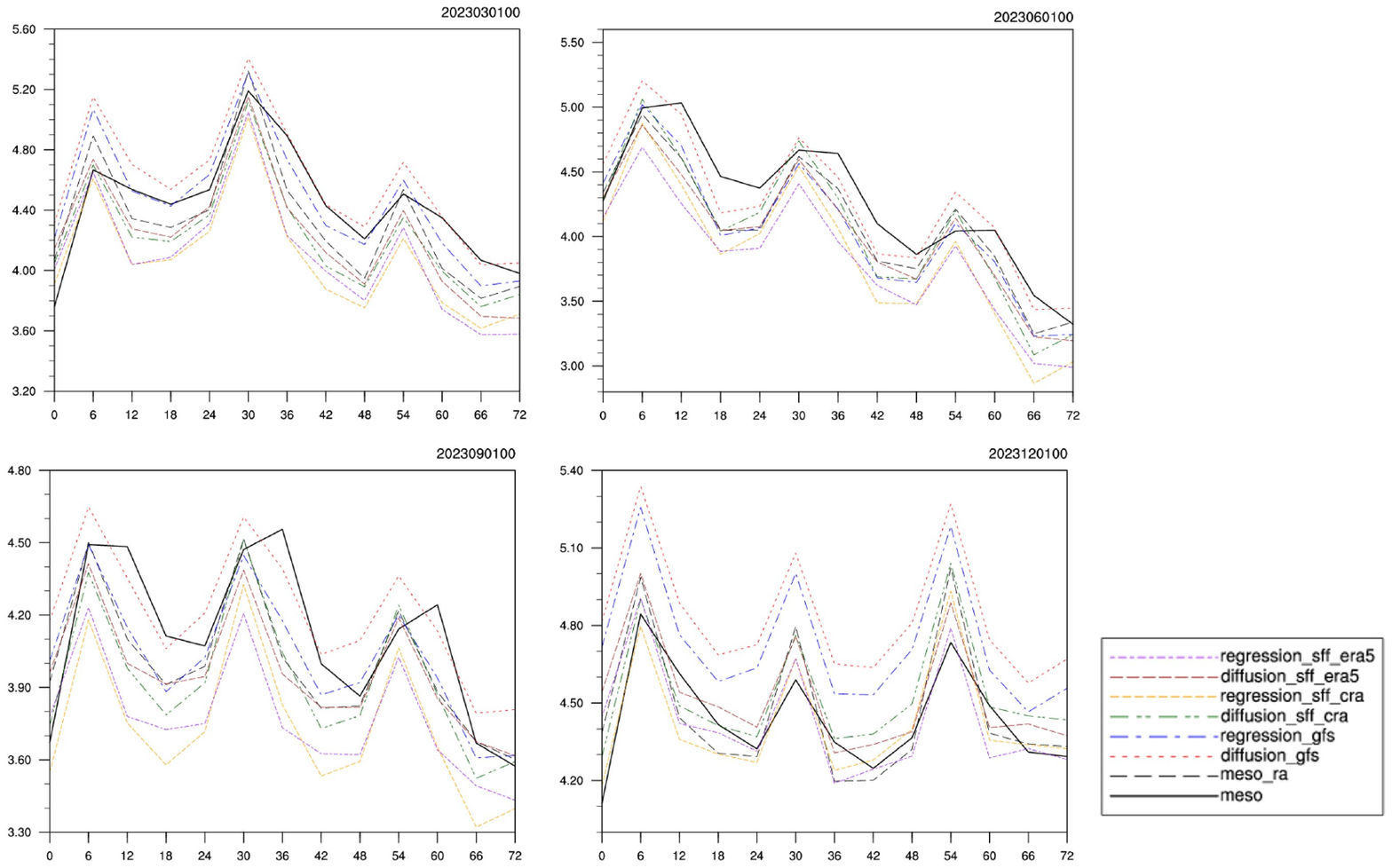}
\caption{Area average of the 10m wind of the 3km reanalysis data and the 3km forecasts of CMA-MESO, downscaled SFF (taking CRA1.5 or ERA5 as initial fields), and downscaled CMA-GFS. Four initialization times are considered: 2023-03-01-00, 2023-06-01-00, 2023-09-01-00, and 2023-12-01-00 UTC. "meso\_ra" denotes the 3km reanalysis; "regression" and "diffusion" refer to the Regression 4 and CorrDiff 4, respectively.
} \label{fig_area_ave_10mwind}
\end{figure}

Fig. \ref{fig_area_ave_2mt} and Fig. \ref{fig_area_ave_10mwind} show the area average of the 3km reanalysis data and 3km forecasts for the 2m temperature and 10m wind.
The forecasts of the 2m temperature are fairly consistent with the 3km reanalysis data, but there are significant inter-model differences for the 10m wind.
The 10m wind trend in the 3km reanalysis data clearly differs from that in the CMA-MESO forecasts. 
All forecasts based on the downscaling models exhibit trends more closely aligned with the 3km reanalysis data. Among them, the downscaled CMA-GFS forecasts produce higher wind speeds than the downscaled SFF forecasts. 
This difference is likely due to the over-smoothing of the outputs of data-driven models like SFF, which reduces intensity.

\subsubsection{Case Study: Tropical Cyclone}
\begin{figure}[!htbp]
\centering
\includegraphics[width=1\linewidth]{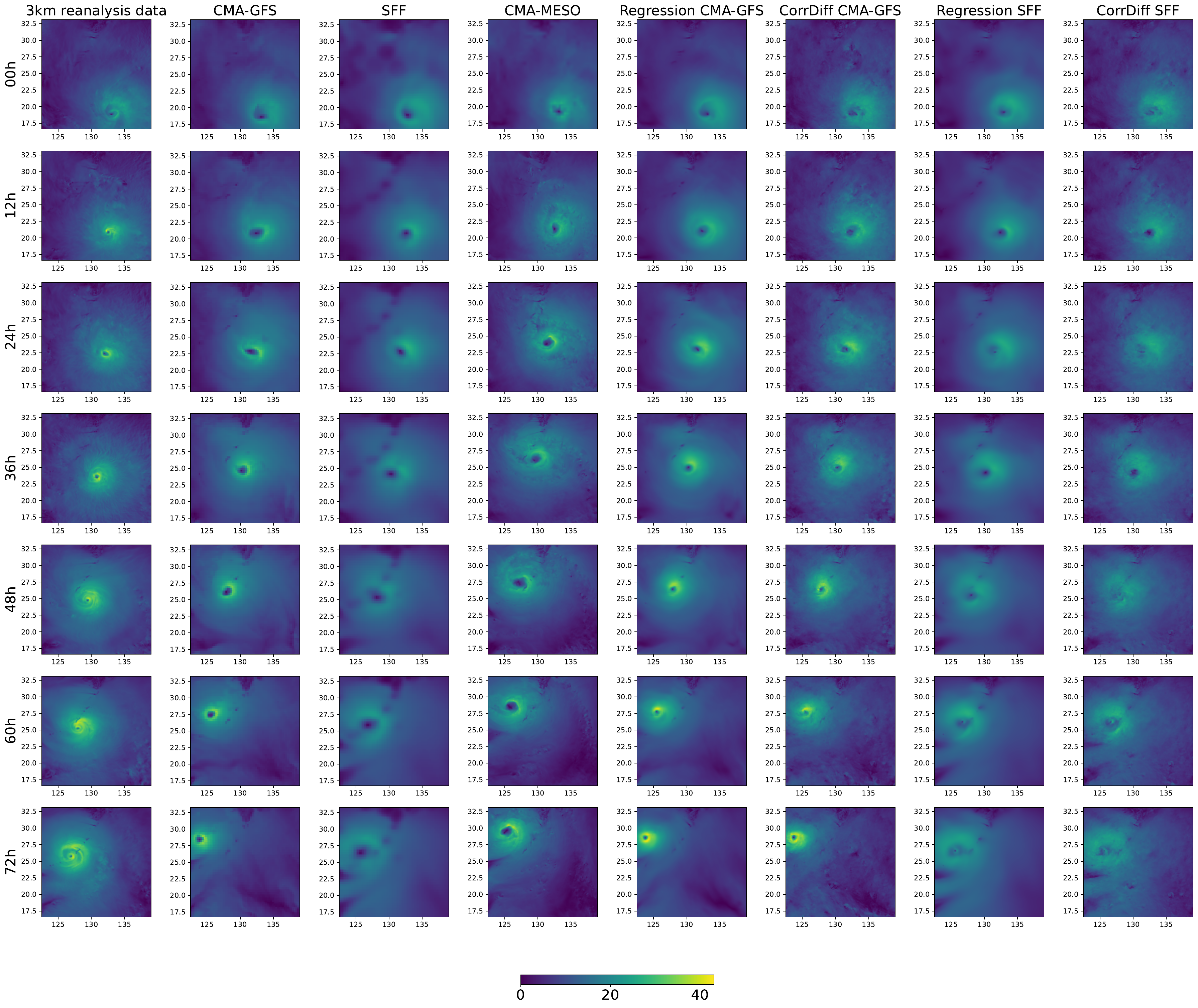}
\caption{Illustration of the downscaling of CMA-GFS and SFF forecasts of Typhoon Khanun by Regression 4 and CorrDiff 4, compared to the 3km reanalysis data, CMA-GFS, SFF and CMA-MESO. Figures represent 10m wind speed. The initialization time is 2023-07-30-00 UTC. The initial fields of SFF are CRA1.5.} \label{fig_72h_Khanun}
\end{figure}

Fig. \ref{fig_72h_Khanun} shows the 10m wind speed of the downscaling of Typhoon Khanun by Regression 4 and CorrDiff 4. The downscaling models are applied to the forecasts of CMA-GFS and SFF that use CRA1.5 as initial fields.
Generally, the outputs of the regression models are smoother and less sharp than those of CMA-MESO, which should be due to the over-smoothing of data-driven models.
Consequently, the Regression SFF forecasts, which involves two data-driven models (SFF and Regression 4), exhibit the most smoothness. The use of diffusion model (CorrDiff 4) notably reduces this effect.
Because of the blurriness, the fine-scale structure of the typhoon cannot be clearly captured by data-driven models.

For the last row in Fig. \ref{fig_72h_Khanun}, using the 3km reanalysis as the ground truth, the MAE scores of CMA-MESO, Regression CMA-GFS, and Regression SFF are 5.24 m/s, 4.82 m/s, and 2.18 m/s, respectively. 
This indicates that a lower MAE does not automatically yield a more realistic visual representation of typhoon winds. 
Note that the Khanun storm center location of SFF forecasts is closer to that of the 3km reanalysis data than that of CMA-GFS and CMA-MESO, which should be a reason for the lower MAE scores.
In fact, typically data-driven models outperform numerical weather prediction models in the inference of the path of a typhoon, as data-driven models are usually better at predicting large-scale patterns.
In order to improve the visual quality of the 10m wind speed forecasts for typhoons, designing a more sophisticated loss function could be a promising direction.
Although, for these results, data-driven downscaling models do not generate visually realistic results of the 10m wind, their inference of the typhoon eye radius might be better than CMA-MESO, indicating that such data-driven models have the potential to refine the typhoon structure in coarse-resolution forecasts.

\begin{figure}[!htbp]
\centering
\includegraphics[width=0.8\linewidth]{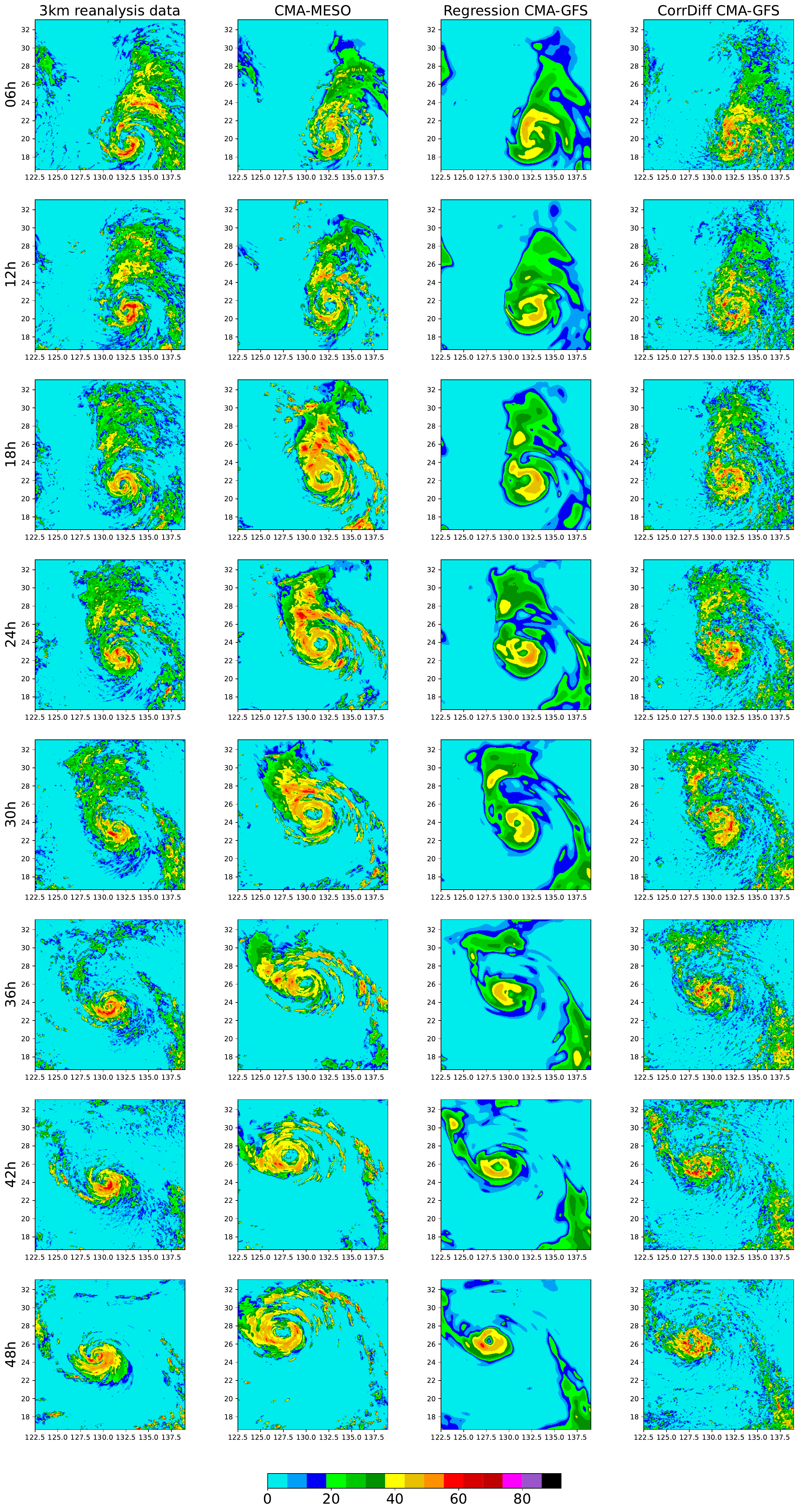}
\caption{Illustration of the 3km radar composite reflectivity forecasts of Typhoon Khanun by applying Regression 2 and CorrDiff 2 on the CMA-GFS forecasts. The initialization time is 2023-07-30-00 UTC.} \label{fig_72h_radar_Khanun}
\end{figure}

Fig. \ref{fig_72h_radar_Khanun} shows the 3km radar composite reflectivity forecasts for Typhoon Khanun, obtained by applying Regression 2 and CorrDiff 2 to the CMA-GFS forecasts. 
The regression model exhibits overly smooth results, failing to predict high radar reflectivity, and it cannot reproduce small reflectivity cores, indicating the limitation of regression models to predict the fine-scale extreme convective precipitation.
By contract, the CorrDiff model exhibits a more realistic spatial distribution.
The high-frequency features of the CorrDiff model are closer to those of the 3km reanalysis data, compared to CMA-MESO.

\begin{figure}[!htbp]
\centering
\includegraphics[width=1\linewidth]{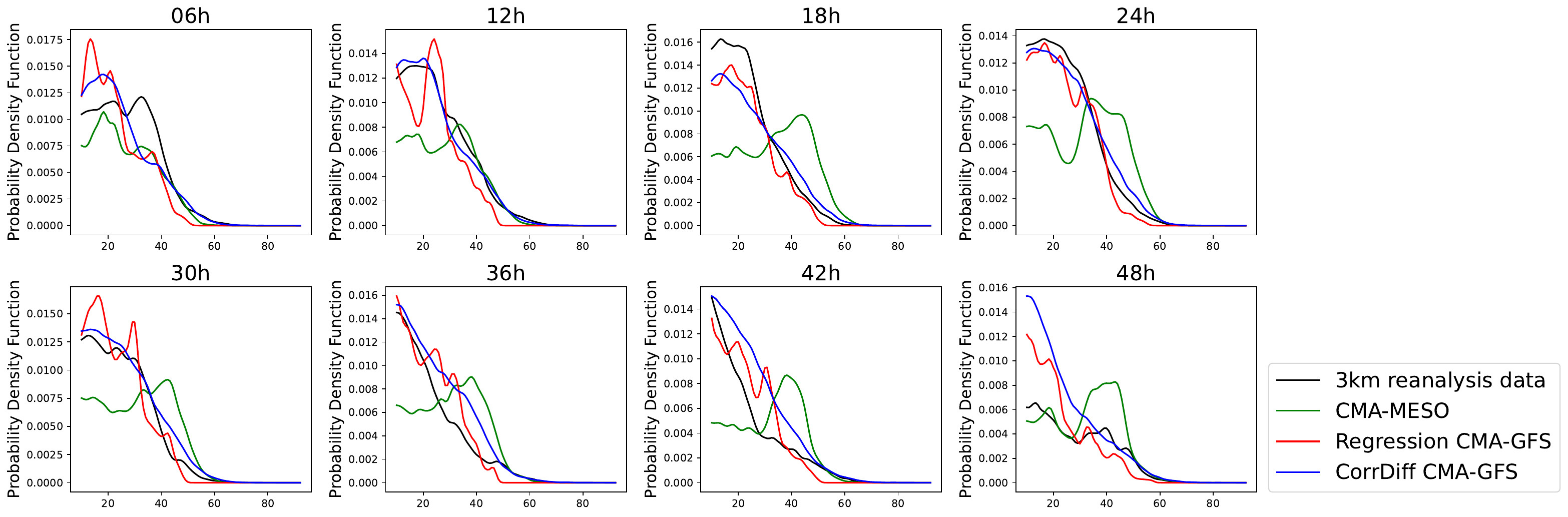}
\caption{Probability density functions of the 3km radar composite reflectivity forecasts of Typhoon Khanun by applying Regression 2 and CorrDiff 2 on CMA-GFS forecasts, compared with 3km reanlysis data and CMA-MESO. The initialization time is 2023-07-30-00 UTC. Only the probability density of the reflectivity that is higher than 10 dBz is shown for clarity.} \label{fig_radar_Khanun_pdf}
\end{figure}

Fig. \ref{fig_radar_Khanun_pdf} shows the probability density functions (PDFs) for the 3km radar composite reflectivity forecasts of Typhoon Khanun.
Overall, the PDFs of the data-driven models align more closely with those of the 3km reanalysis than with those of CMA-MESO.
This is particularly evident during the first 24 hours for reflectivity exceeding 50 dBZ, where the CorrDiff model (blue curve) more accurately replicates the reanalysis distribution (black curve) compared to CMA-MESO (green curve).
For any forecast time, the regression model (red lines) dramatically underestimates the distribution beyond 50 dBz, which is a consequence of over-smoothing, proving that diffusion-based models outperform deterministic models in the prediction of high radar reflectivity or precipitation.

\begin{figure}[!htbp]
\centering
\includegraphics[width=1\linewidth]{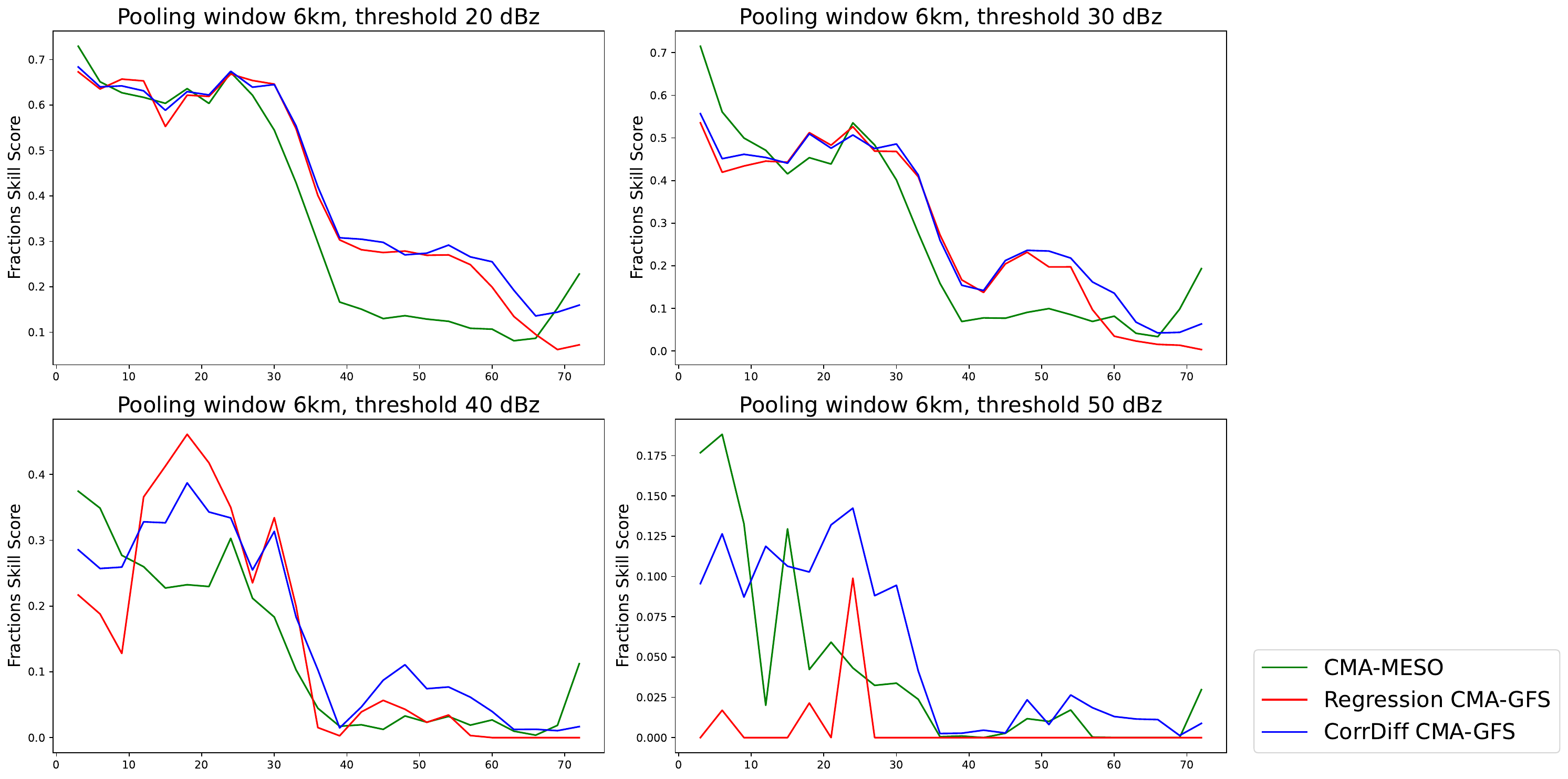}
\caption{Fractions skill scores of the 3km radar composite reflectivity forecasts of Typhoon Khanun by applying Regression 2 and CorrDiff 2 on CMA-GFS forecasts, compared with CMA-MESO, using 3km reanalysis data as ground truth. The initialization time is 2023-07-30-00 UTC.} \label{fig_radar_Khanun_fss}
\end{figure}

The fractions skill scores (FSS) for the 3km radar composite reflectivity forecasts of Typhoon Khanun are given in Fig. \ref{fig_radar_Khanun_fss}.
For all models, there is a dramatic decrease in performance around 36 hours.
In the first 9 hours, CMA-MESO achieves the best FSS.
Between 12 and 36 hours, the CorrDiff model can outperform CMA-MESO.
CorrDiff and regression models have similar FSS for lower reflectivity thresholds, while CorrDiff substantially outperforms the regression model for high thresholds, since the regression model fails to reproduce high reflectivity.

\section{Conclusion}
\label{conclusion}
In this work, we train multiple CorrDiff models with different input/output combinations for 3km downscaling. Evaluation against the CMA-MESO baseline confirms the success of our approach, which generally outperforms CMA-MESO in terms of MAE for the target variables.
However, it is important to note that a lower MAE does not necessarily correspond to better forecasts, particularly for extreme weather events such as typhoons. 
Experimental results also show that, for radar composite reflectivity inference, the CorrDiff model can produce more realistic high-frequency details than the regression model.

One major drawback of our CorrDiff models is that they have higher MAE scores compared to the regression models.
This phenomenon was also observed in the original work of CorrDiff. However, in our experiments, the difference in the MAE scores between CorrDiff and regression models is more obvious, which might be due to the fact that the size of our high-resolution grid is much larger than that in the original work. 
Another limitation is the inference speed of CorrDiff models. Owing to the iterative denoising process, the inference time for a diffusion model is multiple times longer than that of a regression model. 
This computational overhead increases considerably when generating an ensemble of downscaling results.

Future work could explore several promising directions:
\begin{itemize}
\item In current work, only reanalysis data are used to train the models, in order to improve model accuracy, pre-trained models can be further finetuned on operational data. Moreover, they can be finetuned together with SFF.

\item Our demonstration of a correlation between the predictive uncertainty quantified by CorrDiff and the MAE enables the future development of methods that use uncertainty estimates to mitigate forecast errors.

\item A challenge remains in the interpretability of deep learning models \cite{zhang2021survey}. Various methods (for example, gradient-based approaches \cite{simonyan2013deep,sundararajan2017axiomatic}) have been developed for computer vision tasks such as image classification. These techniques could be adapted to elucidate the predictions of CorrDiff by incorporating physical principles.

\end{itemize}

\section*{Acknowledgements}
This work was supported by the National Natural Science Foundation of China (Grant No. U2242210 and No. U2342220). 
We would like to thank Zhifang Xu and Jilin Wang for offering the 3km reanalysis data.
We thank Li Zhang and Siyuan Sun for the evaluation of our 3km forecasts.
We thank Simeng Qian and Chenyu Wang for providing the forecasts of SFF.
We also thank Zhiyan Jin, Huadong Xiao, Qilong Jia and Tongda Xu for the fruitful discussions.

\bibliographystyle{splncs04}
\bibliography{ref}

@article{liping2022key,
  title={Key technologies of CMA-MESO and application to operational forecast},
  author={Liping, Huang and Liantang, Deng and Ruichun, Wang and Zhaorong, Zhuang and Yuan, Jiang and Zhifang, Xu and Lijuan, Zhu and Jin, Zhang and Lili, Wang and Fei, Yu and others},
  journal={Journal of Applied Meteorological Science},
  volume={33},
  number={6},
  pages={641--654},
  year={2022}
}

@article{chen2008new,
  title={New generation of multi-scale NWP system (GRAPES): General scientific design},
  author={Chen, DeHui and Xue, JiShan and Yang, XueSheng and Zhang, HongLiang and Shen, XueShun and Hu, JiangLin and Wang, Yu and Ji, LiRen and Chen, JiaBin},
  journal={Chinese Science Bulletin},
  volume={53},
  number={22},
  pages={3433--3445},
  year={2008},
  publisher={Springer}
}

@article{schoof2001downscaling,
  title={Downscaling temperature and precipitation: A comparison of regression-based methods and artificial neural networks},
  author={Schoof, Justin T and Pryor, SC},
  journal={International Journal of Climatology: A Journal of the Royal Meteorological Society},
  volume={21},
  number={7},
  pages={773--790},
  year={2001},
  publisher={John Wiley \& Sons, Ltd. Chichester, UK}
}

@article{chen2010downscaling,
  title={Downscaling GCMs using the Smooth Support Vector Machine method to predict daily precipitation in the Hanjiang Basin},
  author={Chen, Hua and Guo, Jing and Xiong, Wei and Guo, Shenglian and Xu, Chong-Yu},
  journal={Advances in Atmospheric Sciences},
  volume={27},
  pages={274--284},
  year={2010},
  publisher={Springer}
}

@article{davy2010statistical,
  title={Statistical downscaling of wind variability from meteorological fields},
  author={Davy, Robert J and Woods, Milton J and Russell, Christopher J and Coppin, Peter A},
  journal={Boundary-layer meteorology},
  volume={135},
  pages={161--175},
  year={2010},
  publisher={Springer}
}

@article{laddimath2019artificial,
  title={Artificial neural network technique for statistical downscaling of global climate model},
  author={Laddimath, Rajashekhar S and Patil, Nagraj S},
  journal={Mapan},
  volume={34},
  number={1},
  pages={121--127},
  year={2019},
  publisher={Springer}
}

@article{watt2024generative,
  title={Generative diffusion-based downscaling for climate},
  author={Watt, Robbie A and Mansfield, Laura A},
  journal={arXiv preprint arXiv:2404.17752},
  year={2024}
}

@article{addison2022machine,
  title={Machine learning emulation of a local-scale UK climate model},
  author={Addison, Henry and Kendon, Elizabeth and Ravuri, Suman and Aitchison, Laurence and Watson, Peter AG},
  journal={arXiv preprint arXiv:2211.16116},
  year={2022}
}

@article{mardani2025residual,
  title={Residual corrective diffusion modeling for km-scale atmospheric downscaling},
  author={Mardani, Morteza and Brenowitz, Noah and Cohen, Yair and Pathak, Jaideep and Chen, Chieh-Yu and Liu, Cheng-Chin and Vahdat, Arash and Nabian, Mohammad Amin and Ge, Tao and Subramaniam, Akshay and others},
  journal={Communications Earth \& Environment},
  volume={6},
  number={1},
  pages={124},
  year={2025},
  publisher={Nature Publishing Group UK London}
}

@article{sun2024deep,
  title={Deep learning in statistical downscaling for deriving high spatial resolution gridded meteorological data: A systematic review},
  author={Sun, Yongjian and Deng, Kefeng and Ren, Kaijun and Liu, Jia and Deng, Chongjiu and Jin, Yongjun},
  journal={ISPRS Journal of Photogrammetry and Remote Sensing},
  volume={208},
  pages={14--38},
  year={2024},
  publisher={Elsevier}
}

@article{karras2022elucidating,
  title={Elucidating the design space of diffusion-based generative models},
  author={Karras, Tero and Aittala, Miika and Aila, Timo and Laine, Samuli},
  journal={Advances in neural information processing systems},
  volume={35},
  pages={26565--26577},
  year={2022}
}

@article{ho2020denoising,
  title={Denoising diffusion probabilistic models},
  author={Ho, Jonathan and Jain, Ajay and Abbeel, Pieter},
  journal={Advances in neural information processing systems},
  volume={33},
  pages={6840--6851},
  year={2020}
}

@article{zhang2021survey,
  title={A survey on neural network interpretability},
  author={Zhang, Yu and Ti{\v{n}}o, Peter and Leonardis, Ale{\v{s}} and Tang, Ke},
  journal={IEEE transactions on emerging topics in computational intelligence},
  volume={5},
  number={5},
  pages={726--742},
  year={2021},
  publisher={IEEE}
}

@article{simonyan2013deep,
  title={Deep inside convolutional networks: Visualising image classification models and saliency maps},
  author={Simonyan, Karen and Vedaldi, Andrea and Zisserman, Andrew},
  journal={arXiv preprint arXiv:1312.6034},
  year={2013}
}

@inproceedings{sundararajan2017axiomatic,
  title={Axiomatic attribution for deep networks},
  author={Sundararajan, Mukund and Taly, Ankur and Yan, Qiqi},
  booktitle={International conference on machine learning},
  pages={3319--3328},
  year={2017},
  organization={PMLR}
}

@article{bi2022pangu,
  title={Pangu-weather: A 3d high-resolution model for fast and accurate global weather forecast},
  author={Bi, Kaifeng and Xie, Lingxi and Zhang, Hengheng and Chen, Xin and Gu, Xiaotao and Tian, Qi},
  journal={arXiv preprint arXiv:2211.02556},
  year={2022}
}

@article{lam2023learning,
  title={Learning skillful medium-range global weather forecasting},
  author={Lam, Remi and Sanchez-Gonzalez, Alvaro and Willson, Matthew and Wirnsberger, Peter and Fortunato, Meire and Alet, Ferran and Ravuri, Suman and Ewalds, Timo and Eaton-Rosen, Zach and Hu, Weihua and others},
  journal={Science},
  volume={382},
  number={6677},
  pages={1416--1421},
  year={2023},
  publisher={American Association for the Advancement of Science}
}

@article{pathak2022fourcastnet,
  title={Fourcastnet: A global data-driven high-resolution weather model using adaptive fourier neural operators},
  author={Pathak, Jaideep and Subramanian, Shashank and Harrington, Peter and Raja, Sanjeev and Chattopadhyay, Ashesh and Mardani, Morteza and Kurth, Thorsten and Hall, David and Li, Zongyi and Azizzadenesheli, Kamyar and others},
  journal={arXiv preprint arXiv:2202.11214},
  year={2022}
}

@article{wu2024gsdnet,
  title={GSDNet: A deep learning model for downscaling the significant wave height based on NAFNet},
  author={Wu, Xiaoyu and Zhao, Rui and Chen, Hongyi and Wang, Zijia and Yu, Chen and Jiang, Xingjie and Liu, Weiguo and Song, Zhenya},
  journal={Journal of Sea Research},
  volume={198},
  pages={102482},
  year={2024},
  publisher={Elsevier}
}

@article{chen2023fuxi,
  title={FuXi: a cascade machine learning forecasting system for 15-day global weather forecast},
  author={Chen, Lei and Zhong, Xiaohui and Zhang, Feng and Cheng, Yuan and Xu, Yinghui and Qi, Yuan and Li, Hao},
  journal={npj climate and atmospheric science},
  volume={6},
  number={1},
  pages={190},
  year={2023},
  publisher={Nature Publishing Group UK London}
}

@article{hersbach2000decomposition,
  title={Decomposition of the continuous ranked probability score for ensemble prediction systems},
  author={Hersbach, Hans},
  journal={Weather and Forecasting},
  volume={15},
  number={5},
  pages={559--570},
  year={2000}
}

@inproceedings{bonev2023spherical,
  title={Spherical fourier neural operators: Learning stable dynamics on the sphere},
  author={Bonev, Boris and Kurth, Thorsten and Hundt, Christian and Pathak, Jaideep and Baust, Maximilian and Kashinath, Karthik and Anandkumar, Anima},
  booktitle={International conference on machine learning},
  pages={2806--2823},
  year={2023},
  organization={PMLR}
}

@article{hersbach2020era5,
  title={The ERA5 global reanalysis},
  author={Hersbach, Hans and Bell, Bill and Berrisford, Paul and Hirahara, Shoji and Hor{\'a}nyi, Andr{\'a}s and Mu{\~n}oz-Sabater, Joaqu{\'\i}n and Nicolas, Julien and Peubey, Carole and Radu, Raluca and Schepers, Dinand and others},
  journal={Quarterly journal of the royal meteorological society},
  volume={146},
  number={730},
  pages={1999--2049},
  year={2020},
  publisher={Wiley Online Library}
}

\end{document}